\documentclass{article}

\PassOptionsToPackage{numbers, compress}{natbib}

 \usepackage[eandd,final]{neurips_2026}

\usepackage[utf8]{inputenc} % allow utf-8 input
\usepackage[T1]{fontenc}    % use 8-bit T1 fonts
\usepackage{float}          % must be loaded before hyperref
\usepackage{hyperref}       % hyperlinks
\usepackage{url}            % simple URL typesetting
\usepackage{booktabs}       % professional-quality tables
\usepackage{amsfonts}       % blackboard math symbols
\usepackage{nicefrac}       % compact symbols for 1/2, etc.
\usepackage{microtype}      % microtypography
\usepackage{xcolor}         % colors
\usepackage{xspace}
\usepackage{graphicx}
\usepackage{amsmath}
\usepackage{cleveref}
\usepackage{multirow}
\usepackage{adjustbox}
\usepackage{listings}
\usepackage{enumitem}

\crefname{lstlisting}{Listing}{listings}

\crefname{figure}{Figure}{Figure}
\Crefname{figure}{Figure}{Figure}
\crefname{table}{Table}{Table}
\Crefname{table}{Table}{Table}
\crefname{section}{Section}{Section}
\Crefname{section}{Section}{Section}
\crefname{subsection}{Section}{Section}
\Crefname{subsection}{Section}{Section}
\crefname{subsubsection}{Section}{Section}
\Crefname{subsubsection}{Section}{Section}
\crefname{appendix}{Appendix}{Appendix}
\Crefname{appendix}{Appendix}{Appendix}
\crefname{equation}{Equation}{Equation}
\Crefname{equation}{Equation}{Equation}
\crefname{algorithm}{Algorithm}{Algorithm}
\Crefname{algorithm}{Algorithm}{Algorithm}

\title{CrafterDojo: An Integrated Research Suite\\for Building Open-Ended Embodied Agents in Crafter}

\author{%
  Junyeong Park\thanks{Corresponding authors: \texttt{jyp10987@kaist.ac.kr}, \texttt{sungjin.ahn@kaist.ac.kr}} \\
  KAIST\\
  \texttt{jyp10987@kaist.ac.kr}
  \And
  Hyeonseo Cho \\
  KAIST\\
  \texttt{hyeonseo@kaist.ac.kr}
  \And
  Sungjin Ahn\footnotemark[1] \\
  KAIST\\
  \texttt{sungjin.ahn@kaist.ac.kr}
}

\newcommand{\cvpt}{C-VPT\xspace}
\newcommand{\csteveone}{C-Steve-1\xspace}
\newcommand{\cclip}{C-CLIP\xspace}
\newcommand{\ours}{CrafterDojo\xspace}

\newcommand{\hdep}{H-DEP\xspace}
\newcommand{\dep}{DEP\xspace}

\newcommand{\playdata}{CrafterPlay\xspace}
\newcommand{\clipcaption}{CrafterCaption\xspace}

\newcommand{\crafterwood}{\raisebox{-0.3ex}{\includegraphics[width=10px]{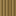}}\xspace}

\newcommand{\crafterwoodpickaxe}{\raisebox{-0.3ex}{\includegraphics[width=10px]{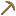}}\xspace}
\newcommand{\crafterwoodsword}{\raisebox{-0.3ex}{\includegraphics[width=10px]{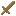}}\xspace}

\newcommand{\crafteriron}{\raisebox{-0.3ex}{\includegraphics[width=10px]{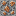}}\xspace}

\newcommand{\craftertable}{\raisebox{-0.3ex}{\includegraphics[width=10px]{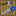}}\xspace}

\newcommand{\crafterstone}{\raisebox{-0.3ex}{\includegraphics[width=10px]{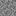}}\xspace}
\newcommand{\crafterdiamond}{\raisebox{-0.3ex}{\includegraphics[width=10px]{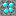}}\xspace}
\newcommand{\crafterplayer}{\raisebox{-0.3ex}{\includegraphics[width=10px]{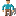}}\xspace}

\newcommand{\crafterdrink}{\raisebox{-0.3ex}{\includegraphics[width=10px]{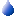}}\xspace}
\newcommand{\crafterenergy}{\raisebox{-0.3ex}{\includegraphics[width=10px]{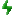}}\xspace}
\newcommand{\crafterfood}{\raisebox{-0.3ex}{\includegraphics[width=10px]{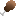}}\xspace}
\newcommand{\crafterzombie}{\raisebox{-0.3ex}{\includegraphics[width=10px]{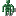}}\xspace}
\newcommand{\crafterskeleton}{\raisebox{-0.3ex}{\includegraphics[width=10px]{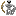}}\xspace}

\newcommand{\minecraftlog}{\raisebox{-0.3ex}{\includegraphics[width=10px]{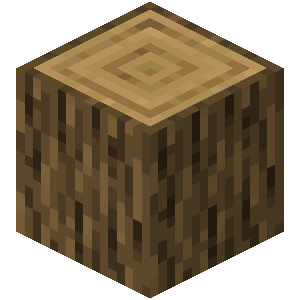}}\xspace}
\newcommand{\minecraftstone}{\raisebox{-0.3ex}{\includegraphics[width=10px]{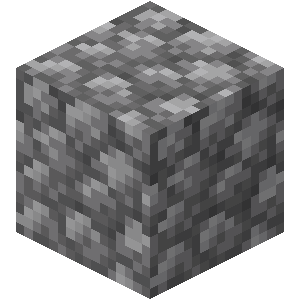}}\xspace}
\newcommand{\minecraftsironore}{\raisebox{-0.3ex}{\includegraphics[width=10px]{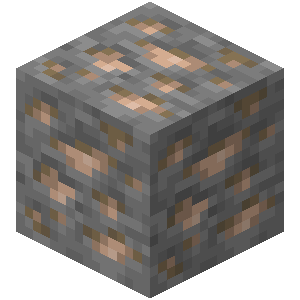}}\xspace}
\newcommand{\minecraftdiamond}{\raisebox{-0.3ex}{\includegraphics[width=10px]{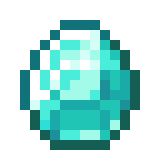}}\xspace}

\definecolor{codegreen}{rgb}{0,0.6,0}
\definecolor{codegray}{rgb}{0.5,0.5,0.5}
\definecolor{codepurple}{rgb}{0.58,0,0.82}
\definecolor{backcolour}{rgb}{0.95,0.95,0.92}

\lstdefinestyle{mystyle}{
    backgroundcolor=\color{backcolour},   
    commentstyle=\color{codegreen},
    keywordstyle=\color{magenta},
    numberstyle=\tiny\color{codegray},
    stringstyle=\color{codepurple},
    basicstyle=\ttfamily\scriptsize,
    breakatwhitespace=false,         
    breaklines=true,                 
    captionpos=b,                    
    keepspaces=true,                 
    numbers=none,                    
    numbersep=5pt,                  
    showspaces=false,                
    showstringspaces=false,
    showtabs=false,                  
    tabsize=2,
    frame=none,
    aboveskip=1pt,
    belowskip=1pt,
}
\begin{document}

\maketitle

\begin{abstract}
Foundation models are central to building general-purpose embodied agents. Minecraft has become the dominant testbed for this paradigm, but its computational and engineering overhead limits rapid iteration and controlled analysis. Crafter offers a lightweight alternative yet lacks the foundation model ecosystem required for such research. We introduce \ours, a suite of foundation models, datasets, and data-generation toolkits that establishes Crafter as a testbed for foundation-model-based embodied intelligence, providing \emph{CrafterVPT}, \emph{CrafterCLIP}, and \emph{CrafterSteve-1}. To address Crafter's data scarcity, \ours includes automatic tools that generate diverse synthetic demonstrations and aligned video-language pairs, yielding the \emph{CrafterPlay} and \emph{CrafterCaption} datasets. Experiments show that these datasets support effective foundation model training and enable hierarchical agents that outperform non-hierarchical baselines on long-horizon tasks. We release all code, models, and datasets at \url{https://github.com/frechele/CrafterDojo}.
\end{abstract}

\section{Introduction}

Developing general-purpose embodied agents that follow diverse instructions, plan over long horizons, and adapt to open-ended environments remains a central challenge in AI~\citep{guss_minerl_2019,fan_minedojo_2022,hafner2025training}. 
Pure reinforcement learning struggles in such settings: sparse rewards and vast search spaces hinder exploration, while task diversity makes independent learning impractical. 
Consequently, embodied agent research has increasingly adopted foundation models, which provide behavioral priors, commonsense knowledge, and reasoning capabilities composable into more capable systems~\citep{ahn2022can, wang_jarvis-1_2023, cai_groot_2024, park_mrsteve_2025}.

Minecraft~\citep{guss_minerl_2019, fan_minedojo_2022} has emerged as the dominant testbed for this paradigm because it supports a rich ecosystem of foundation models trained on large-scale internet data. 
These models provide behavioral priors, vision-language grounding, and instruction-following capabilities, enabling hierarchical and modular agents that combine high-level planning with learned low-level control~\citep{wang_jarvis-1_2023, yuan_pre-training_2024}. 
However, Minecraft poses substantial practical barriers: its simulation is slow, resource-intensive, and crash-prone~\citep{nottingham_embodied_2023}. 
Moreover, its 3D visuals and continuous control often conflate multiple failure modes---perception, low-level control, navigation, and planning---making it difficult to isolate algorithmic improvements. 
Consequently, the cost and fragility of experimentation in Minecraft impede rapid iteration, limiting controlled studies and slowing the pace of scientific discovery.

Crafter~\citep{hafner_benchmarking_2022} offers an appealing alternative. 
It preserves the essential structure of open-ended embodied tasks---procedurally generated worlds, sparse rewards, and a technology tree requiring long-horizon planning---while removing confounding complexities from 3D perception and continuous control. 
As a lightweight 2D gridworld implemented in pure Python, Crafter enables fast simulation, easy modification, and systematic ablations, making it well suited for controlled studies of planning, memory, and exploration. 
Crafter thus provides a cleaner environment for analyzing the mechanisms of embodied intelligence. Its simulation also runs significantly faster than Minecraft (\cref{sec:environment_speed_comp}) while preserving comparable prerequisite structure (\cref{sec:task_difficulty_comp}).

\begin{figure*}[t]
    \centering
    \includegraphics[width=0.95\textwidth]{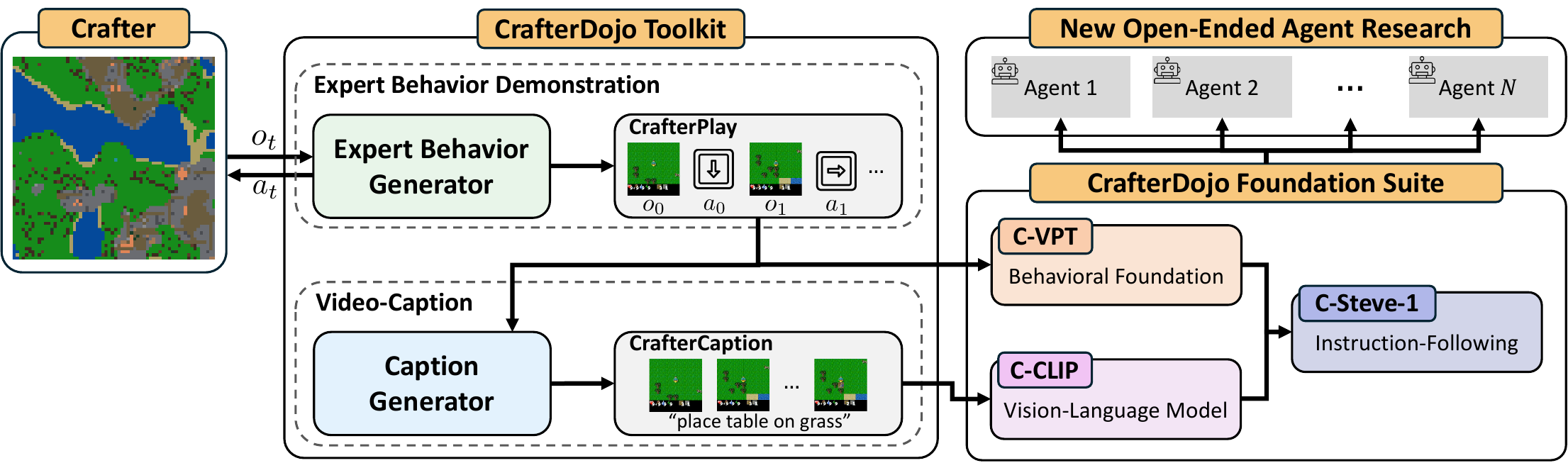}
    \caption{Overview of \ours components. \ours provides two automatic dataset generation toolkits (Expert Behavior Generator and Caption Generator) that automatically generate large-scale datasets to train three foundation models: \cvpt for behavioral foundation, \cclip for vision-language grounding, and \csteveone for instruction-following in Crafter. With these models, datasets, and toolkits, researchers can efficiently develop new agents without the overhead of working in the heavy and complex Minecraft environment.}
    \label{fig:craftdojo_components}
\end{figure*}

Despite these advantages and recent uses of Crafter and its derivatives~\citep{zhou2025walle,fu2026counterfactual,khan2026onelife}, the research community has not been able to fully benefit from Crafter for foundation-model-based embodied agent research.
Unlike Minecraft's abundant gameplay videos, tutorials, and wiki pages, Crafter lacks the large-scale demonstrations and language annotations required to pre-train foundation models.
As a result, existing work has either relied on end-to-end reinforcement learning~\citep{du_guiding_2023, hafner_mastering_2023, micheli_efficient_2024, dedieu_improving_2025}, which struggles with sparse rewards and vast search spaces inherent to long-horizon tasks, or on LLM-based agents that invoke language models at every timestep for primitive action selection~\citep{wu_spring_2023, sun_enhancing_2024}, incurring substantial inference costs and limiting scalability.

Our goal is not to propose a new agent architecture, but to establish Crafter as a first-class testbed for foundation-model-based embodied intelligence, facilitating rapid iteration and accelerating scientific progress in this field.
To this end, we introduce \textbf{\ours}, a suite of foundation models, datasets, and toolkits that brings to Crafter the same research paradigm enabled by VPT~\citep{baker_video_2022}, MineCLIP~\citep{fan_minedojo_2022}, and Steve-1~\citep{lifshitz_steve-1_2023} in Minecraft. 
Just as these models have served as foundations for diverse agent research in Minecraft~\citep{wang_jarvis-1_2023, zhou_minedreamer_2024, yuan_pre-training_2024, park_mrsteve_2025}, \ours provides the analogous building blocks for Crafter: \emph{CrafterVPT} for behavioral priors, \emph{CrafterCLIP} for vision-language grounding, and \emph{CrafterSteve-1} for instruction-following. 
By combining these pre-trained models with Crafter's fast simulation, researchers can complete the full research loop---pretraining, evaluation, ablation, and debugging---in a fraction of the time required in Minecraft. \ours is built on Craftax-Classic~\citep{matthews_craftax_2024}, a fast JAX reimplementation of Crafter with closely matched mechanics and the same 22 achievements, enabling large-scale expert training and data generation infeasible in the original Crafter.

A central challenge in building such a stack is the absence of training data in Crafter. 
To address this, \ours includes two automatic data-generation toolkits. The \emph{Expert Behavior Generator} produces large-scale synthetic demonstrations by training expert policies and rolling them out in Craftax-Classic, yielding the \emph{CrafterPlay} dataset for behavioral pretraining. The \emph{Caption Generator} creates aligned video--language pairs using rule-based event detection and LLM-based paraphrasing, producing the \emph{CrafterCaption} dataset for vision-language grounding. Both toolkits are modular, letting researchers customize reward functions, caption rules, and data distributions.

Our main contributions are threefold:
(i) \textbf{Foundation models}---CrafterVPT, CrafterCLIP, and CrafterSteve-1---that together form the first reusable foundation stack for Crafter, enabling foundation-model-based embodied agents to be studied in a controlled setting;
(ii) \textbf{Datasets and toolkits}---CrafterPlay, CrafterCaption, and automatic data generation tools that address Crafter’s absence of training data and support fast, low-cost iteration; and
(iii) \textbf{Reference agents}---implementations demonstrating how these models can be composed into hierarchical agents. \cref{fig:craftdojo_components} provides an overview of \ours and its components.
We release all code, models, datasets, and toolkits.

\section{Related Work}

\textbf{Foundation Models for Embodied Agents.}
Large-scale pre-training has transformed embodied agents by learning rich behavioral priors and multi-modal representations from extensive demonstration data. These foundation models can be reused through fine-tuning, prompting, or integration into hierarchical systems. This paradigm has proven effective across robotics~\citep{brohan_rt-1_2022, zitkovich_rt-2_2023, octo_model_team_octo_2024} and simulation ~\citep{reed_generalist_2022, wang_jarvis-1_2023}, where large-scale pre-training enables generalization to new open-ended environments.

Minecraft is a leading testbed for open-ended embodied agents, offering procedural world generation, diverse long-horizon tasks, and large-scale data~\citep{guss_minerl_2019, fan_minedojo_2022}. Key foundation models---VPT for behavioral priors~\citep{baker_video_2022}, MineCLIP for vision-language grounding~\citep{fan_minedojo_2022}, and Steve-1 for instruction-following~\citep{lifshitz_steve-1_2023}---have become building blocks for increasingly capable agents~\citep{wang_describe_2023, wang_jarvis-1_2023, zhou_minedreamer_2024, qin_mp5_2024, li_optimus-1_2024,park_mrsteve_2025}.

However, Minecraft's significant computational complexity and engineering overhead limit fast ideation and validation of ideas for open-ended agents.

\textbf{Crafter and Foundation Model Gap.} Crafter~\citep{hafner_benchmarking_2022} addresses Minecraft's limitations with a lightweight 2D benchmark preserving core challenges: procedural generation, crafting, and survival. Recent work includes various agents~\citep{du_guiding_2023, wu_spring_2023, micheli_efficient_2024, sun_enhancing_2024, dedieu_improving_2025}. Its pure Python implementation and easy modifiability have also enabled novel task sets~\citep{xu_active_2023, tang_mars_2024, stanic_learning_2024}. Despite these advances, Crafter lacks the foundation model ecosystem that enables sophisticated Minecraft research. Without behavioral priors, researchers must train policies from scratch; and without instruction-following agents, downstream research that builds on these components becomes difficult. Our work addresses this gap by providing the first comprehensive foundation model suite for Crafter, enabling the same foundation model-based research paradigm that has driven progress in Minecraft.

\section{\ours: Overview}

\ours establishes Crafter as a practical testbed for foundation-model-based embodied agent research by providing a reusable stack of foundation models and data generation toolkits. 
It mirrors the research ecosystem developed for Minecraft---including behavioral priors, vision-language grounding, and instruction-following agents---while operating in a lightweight and controlled environment.

\ours consists of three foundation models: \textbf{CrafterVPT (\cvpt)} for behavioral priors, \textbf{CrafterCLIP (\cclip)} for vision-language grounding, and \textbf{CrafterSteve-1 (\csteveone)} for instruction following.
Unlike Minecraft, which benefits from abundant human demonstrations, Crafter lacks such resources, requiring data and models to be constructed from scratch.

To support this paradigm, we provide two automatic dataset generation toolkits: the \textbf{Expert Behavior Generator} and the \textbf{Caption Generator}. 
These toolkits offer scalable and controllable sources of behavioral and language supervision, enabling foundation model training without human demonstrations and customization for specific goals.
Specifically, Expert Behavior Generator yields \textbf{\playdata} for behavioral pretraining, and Caption Generator yields \textbf{\clipcaption} for vision-language grounding.

In the following sections, we describe how these components support systematic studies of behavioral priors, vision-language grounding, and instruction-following in Crafter.

\section{\cvpt: Behavioral Foundation}

CrafterVPT (\cvpt) enables the study of behavioral priors in Crafter by adapting the VPT paradigm to a lightweight and controllable environment. 
Whereas Minecraft relies on large-scale human gameplay videos to train such priors, Crafter lacks human demonstrations, making behavioral foundation research infeasible without automatic data generation. 
To address this, we develop the \textbf{Expert Behavior Generator} toolkit, which provides a scalable and controllable source of demonstrations and supports systematic ablations over training data and objectives.

\subsection{Expert Behavior Generator}

\textbf{Toolkit Design.} The Expert Behavior Generator trains expert policies using PPO-RNN in Craftax-Classic-Symbolic~\citep{matthews_craftax_2024}, using symbolic observations to simplify training. \playdata combines trajectories from a Survival and a Gathering expert in equal proportion, each trained for 10B timesteps to all 22 Crafter achievements under a different reward function (full specifications in \cref{sec:playdata_spec}).

\textbf{Dataset Generation.} Using this toolkit, we generated the \playdata dataset $\mathcal{D}_\text{play} = \{ \tau_i \}_{i=1}^{M}$ of $M = 20{,}000$ trajectories (10{,}000 per expert), of which 18{,}000 are used for training, 1{,}000 for validation, and 1{,}000 for testing. Each trajectory $\tau_i = \{ (o_t, a_t) \}_{t=1}^{T_i}$ consists of pixel-based observation-action pairs averaging 7{,}887 steps in length.

\textbf{Toolkit Extensibility.} Beyond the default \playdata, varying the reward shaping or the expert training scale yields qualitatively distinct demonstration distributions, enabling controlled studies of how data composition shapes the behavioral foundation (\cref{sec:cvpt_behavior_ablation,sec:cvpt_expert_scale}).

\subsection{Training Procedure}

\textbf{Model Architecture.} We adopt the identical architecture as VPT~\citep{baker_video_2022}, consisting of a ResNet~\citep{he_deep_2016} that encodes image observations, a TransformerXL that processes the historical context, and a Policy Head that predicts the next action given the historical context:
\begin{equation}
\begin{alignedat}{2}
    \text{Image Encoder:} &\quad& x_t &= \operatorname{ResNet}_\theta(o_t), \\
    \text{TransformerXL:} &\quad& \tilde{x}_{1:t} &= \operatorname{TrXL}_\theta(x_{1:t}), \\
    \text{Policy Head:} &\quad& \hat{a}_t &\sim \pi_\theta(\cdot \mid \tilde{x}_{1:t}).
\end{alignedat}
\end{equation}

We trained two model size variants (tiny and base) from scratch on \playdata, distilling the symbolic expert into a pixel-based behavioral prior (details in \cref{cvpt_spec}).

\textbf{Noop Action Filtering.} \playdata contains about $50\%$ noop (no operation) actions---timesteps where the agent takes no meaningful action or simply waits. The high proportion of inactive behavior can degrade \cvpt performance without filtering. Complete filtering hinders strategic waiting behaviors, such as sheltering from hostile mobs. Thus, we remove noop action sequences shorter than 20 consecutive steps and actions issued while sleeping regardless of sequence length, while otherwise preserving longer noop sequences that may represent strategic waiting, reducing the noop ratio to $3.0\%$ and improving the performance of \cvpt.

\textbf{Training Objective.} Unlike VPT, which requires an Inverse Dynamics Model (IDM) to generate pseudo-labels for unlabeled video demonstrations, \cvpt directly leverages ground truth action labels from our synthetic demonstrations. \cvpt follows the standard VPT training procedure (see \cref{cvpt_spec} for full specifications) with the training objective:
\begin{equation}
\mathcal{L}_\text{cvpt} = \mathbb{E}_{(o_{1:t},\, a_t) \sim \mathcal{D}_\text{play}} \bigl[ -\log \pi_\theta(a_t \mid o_{1:t}) \bigr].
\end{equation}

\section{\cclip: Vision-Language Grounding}
CrafterCLIP (\cclip) enables systematic vision-language grounding research in Crafter by adapting the MineCLIP paradigm~\citep{fan_minedojo_2022} to a lightweight and controllable environment. 
In Minecraft, large-scale video--text pairs from the internet provide natural supervision for grounding. 
In contrast, Crafter lacks such resources, making grounding research infeasible without automatic language annotation.

To make vision-language grounding possible in Crafter, we develop the \textbf{Caption Generator} toolkit. 
This toolkit provides a scalable and extensible source of paired video--language supervision, enabling controlled studies of language grounding and instruction semantics in Crafter.

\subsection{Caption Generator}

\textbf{Toolkit Design.} The Caption Generator toolkit uses a rule-based caption generation system that monitors agent behavior and automatically creates descriptive captions for corresponding video segments. We define 15 rules across four categories: Achievement, Movement, Construct, and Combat. For more details, please refer to \cref{sec:clipcaption_spec}.

\textbf{Rule Matching.} At each time step $t$, the caption generation system checks whether the current action and the resulting state change---including map information and inventory---match any predefined rule. When a match occurs, the system appends the video segment $o_{t-4:t+1}$ along with the corresponding caption to the dataset; the additional frame $o_{t+1}$ is included to capture the outcome of action $a_t$. For instance, if an action $a_{10}$ is ``place stone'' and triggers the ``build shelter'' rule, frames $o_{6:11}$ are added with the caption ``place stone to build shelter''.

\textbf{Dataset Generation.} Applying the Caption Generator to \playdata, we generate the \clipcaption dataset, $\mathcal{D}_\text{cap} = \{ (\mathbf{o}_i, c_i) \}_{i=1}^{M}$, of 6-frame video segments $\mathbf{o}_i$ paired with captions $c_i$. The toolkit yields about 67M raw pairs, which we subsample to a uniform distribution over the 15 caption rules, yielding $M=1.2\text{M}$ pairs that inherit \playdata's split. Details are in \cref{sec:clipcaption_spec}.

\textbf{LLM-based Augmentation.} Template-based captions in \clipcaption suffer from limited linguistic diversity. We therefore add LLM-based augmentation following LaCLIP~\citep{fan_improving_2023}: LLMs generate 40 paraphrases for each of the 61 template-based captions, yielding 2,440 diverse captions. We provide the full prompt and example outputs in \cref{sec:llm_clip_paraphrasing}.

\textbf{Toolkit Extensibility.} The Caption Generator's rule-based system is highly modular and extensible---the toolkit enables easy addition of rules, modification of existing caption templates, or integration of domain-specific vocabulary to generate captions tailored to specific research objectives.

\subsection{Training Procedure}

\textbf{Training Objective.} We initialize the video and text encoders of \cclip from OpenAI CLIP~\citep{radford_learning_2021} and fine-tune on \clipcaption. During training, we sample a batch of $B$ video segments $\{\mathbf{o}_b\}_{b=1}^{B}$ with their original captions $\{c_{b,0}\}_{b=1}^{B}$ from $\mathcal{D}_\text{cap}$. For each $b$, we then uniformly sample paraphrased variants $c^\prime_b \sim \text{Uniform}(c_{b,0}, \dots, c_{b,N})$, where $N$ is the number of caption variants, including the original caption $c_{b,0}$ and $N-1$ paraphrases. These video-caption pairs are used to train \cclip, following the MineCLIP training procedure with the following objective:
\begin{equation}
\mathcal{L}_\text{cclip} = -\sum_{b=1}^{B} \log \frac{\exp\bigl(s(\mathbf{o}_b, c'_b)\bigr)}{\sum_{k=1}^{B} \exp\bigl(s(\mathbf{o}_b, c'_k)\bigr)}, \qquad s(\mathbf{o}, c) = \operatorname{sim}\bigl(E_V(\mathbf{o}), E_T(c)\bigr),
\end{equation}
where $E_V$ denotes the video encoder and $E_T$ denotes the text encoder of \cclip. We report the full training configuration of \cclip in \cref{sec:cclip_spec}.

\section{\csteveone: Instruction-Following Agent} \label{sec:csteveone}
CrafterSteve-1 (\csteveone) enables instruction-following agent research in Crafter by adapting the Steve-1 paradigm~\citep{lifshitz_steve-1_2023} to a lightweight and controllable environment.
Whereas Steve-1 in Minecraft relies on pre-trained behavioral and vision-language foundations (VPT and MineCLIP), the availability of \cvpt and \cclip allows the same framework to be instantiated in Crafter, enabling the study of language-conditioned policies under a foundation-model-based paradigm.

\subsection{Implementation Details}

\textbf{Dataset Generation.} To prepare the dataset for \csteveone, we use trajectories from \playdata and apply packed hindsight relabeling~\citep{lifshitz_steve-1_2023}. However, unlike Minecraft's longer task patterns, Crafter often involves multiple distinct tasks in quick succession (\textit{e.g.}, place table \craftertable, make wood pickaxe \crafterwoodpickaxe, then make wood sword \crafterwoodsword in just 3 steps). Packed hindsight relabeling samples goals at random intervals, which can group these distinct tasks under a single goal, causing suboptimal learning.

To address this, we use event-based packed hindsight relabeling to construct $\mathcal{D}_\text{play}^\text{relabel}$. This approach splits each trajectory at \clipcaption event boundaries into coherent segments of consecutive frames with identical captions. Goals are then sampled uniformly within each segment.

\textbf{Model Architecture.} We modify \cvpt to condition on goals specified by \cclip video embeddings. Specifically, we add affine transformation of a goal embedding $z_{1:t}$ to the TransformerXL output before passing it to the Policy Head:
\begin{equation}
\begin{alignedat}{4}
    \text{Image Encoder:} &\quad& x_t &= \operatorname{ResNet}_\theta(o_t), &\qquad \text{TransformerXL:} &\quad& \tilde{x}_{1:t} &= \operatorname{TrXL}_\theta(x_{1:t}), \\
    \text{Conditioning:} &\quad& \tilde{x}'_{1:t} &= \tilde{x}_{1:t} + W_\theta z_{1:t} + b_\theta, &\qquad \text{Policy Head:} &\quad& \hat{a}_t &\sim \pi_\theta(\cdot \mid \tilde{x}'_{1:t}).
\end{alignedat}
\end{equation}
We note that we conditioned after the TransformerXL, rather than the ResNet as in the original Steve-1, following~\citet{park_mrsteve_2025}, for better performance.

\subsection{Training Procedure}

\textbf{Goal-Conditioned Policy Training.} We apply LoRA~\citep{hu2022lora} to the TransformerXL of \cvpt for efficient fine-tuning while preserving behavioral priors. \csteveone follows goal-conditioned imitation learning with a per-timestep loss weight $w_t$:
\begin{equation}
\mathcal{L}_\text{csteve1} = \mathbb{E}\bigl[ -w_t \log \pi_\theta(a_t \mid o_{1:t}, z_{1:t}) \bigr], \qquad (o_{1:t}, a_t, z_{1:t}) \sim \mathcal{D}_\text{play}^\text{relabel}.
\end{equation}
The weight $w_t \propto \operatorname{clamp}(-\log \pi_\theta(a_t \mid o_{1:t}), 0, c_\text{clamp})$ is normalized to unit mean within each minibatch. Because the same observation can support multiple actions across trajectories, the observation history alone often leaves the next action ambiguous. The weight $w_t$ therefore focuses fine-tuning on examples where the goal-conditioning helps clarify the intended behavior.

\textbf{Prior Model Training.} Following Steve-1, we train a conditional variational autoencoder (CVAE)~\citep{sohn_learning_2015} to translate text instructions into \cclip visual goal embeddings, enabling text-conditioned inference. The CVAE is trained on 117M video-text embedding pairs $\mathcal{D}_\text{prior}$, sampled from 1,000 episodes in \clipcaption, minimizing:
\begin{equation}
\mathcal{L}_\text{prior} = \mathbb{E}_{(z_\text{v},\, z_\text{t}) \sim \mathcal{D}_\text{prior}} \Bigl[ \operatorname{KL}\bigl(q_\phi(z_\text{v} \mid z_\text{t}) \,\|\, p(z_\text{v})\bigr) - \mathbb{E}_{c \sim q_\phi(z_\text{v} \mid z_\text{t})} \bigl[ \log p_\phi(z_\text{v} \mid c, z_\text{t}) \bigr] \Bigr],
\end{equation}
where $z_\text{v}$ and $z_\text{t}$ are the video and text embeddings of a pair sampled from $\mathcal{D}_\text{prior}$.

\textbf{Inference.} During inference, natural language instructions are encoded via \cclip text encoder, mapped to goal embeddings $z_\text{goal}$ using the trained CVAE, then actions are sampled with Classifier-Free Guidance with scale $\lambda$:
\begin{equation}
\text{logits} = (1 + \lambda) \log \pi_\theta(a_t \mid o_{1:t}, z_\text{goal}) - \lambda \log \pi_\theta(a_t \mid o_{1:t}).
\end{equation}
\Cref{sec:csteveone_spec} provides the hyperparameters and further implementation details.

\section{Experiments}

In this section, we evaluate whether \ours establishes Crafter as a practical testbed for foundation-model-based embodied agent research. Specifically, we investigate three research questions that reflect the core requirements:
\begin{enumerate}[leftmargin=*,noitemsep,topsep=0pt]
    \item \textit{RQ1}: Can CrafterDojo's toolkits generate large-scale training data without human demonstrations, enabling foundation model pretraining in Crafter?
    \item \textit{RQ2}: Do the resulting foundation models reliably serve their intended roles (behavioral priors, vision-language grounding, and instruction-following)?
    \item \textit{RQ3}: Can these foundation models be composed into foundation-model-based agents in Crafter, such as hierarchical agents commonly studied in Minecraft?
\end{enumerate}

We address \textit{RQ1} and \textit{RQ2} through component-wise evaluations (\cref{sec:cvpt_eval}, \cref{sec:cclip_eval}, and \cref{sec:csteveone_eval}), and assess \textit{RQ3} using a reference hierarchical agent (\cref{sec:hierarchical_agent_eval}). 
Together, these experiments test whether CrafterDojo provides the essential building blocks for systematic and reusable foundation-model-based agent research in Crafter.

\subsection{Evaluating Behavioral Foundation Model: \cvpt} \label{sec:cvpt_eval}
\begin{table}[t]
\centering
\caption{Crafter Score and Return of baselines and \cvpt at two scales. Environment and Observation give the testbed and input modality. $\dagger$: published values from~\citet{dedieu_improving_2025}.}
\begin{tabular}{ccccc}
    \toprule
    \textbf{Method}                         & \textbf{Environment}    & \textbf{Observation}    & \textbf{Score (\%)}     & \textbf{Return (\%)} \\
    \midrule
    Human $\dagger$                         & Crafter   & Pixel     & $50.5 \pm 6.8$          & $65.0 \pm 10.5$ \\
    Expert Policy                           & Craftax-Classic   & State & $97.3 \pm 1.1$          & $96.1 \pm 0.8$ \\
    \midrule
    $\Delta$-IRIS $\dagger$                 & Crafter   & Pixel     & $9.3$                   & $35.0 \pm 3.2$  \\
    DreamerV3 $\dagger$                     & Crafter   & Pixel     & $14.5 \pm 1.6$          & $53.2 \pm 8.0$  \\
    \citet{dedieu_improving_2025} $\dagger$ & Craftax-Classic   & Pixel     & $31.8 \pm 1.4$          & $69.7 \pm 1.2$ \\
    \midrule
    \cvpt (tiny)                            & Craftax-Classic   & Pixel     & $95.6 \pm 1.6$          & $93.2 \pm 1.4$ \\
    \cvpt (base)                            & Craftax-Classic   & Pixel     & $\mathbf{97.7 \pm 0.7}$ & $\mathbf{95.5 \pm 1.1}$ \\
    \bottomrule
\end{tabular}
\label{tbl:crafter_scores}
\end{table}

We evaluate \cvpt across all 22 Crafter achievements, which cover core skills such as exploration, resource gathering, crafting, and survival. 
This evaluation serves two purposes: (1) assessing whether \playdata enables scalable pretraining without human demonstrations, and (2) testing whether \cvpt can function as a reusable behavioral foundation for Crafter.  To examine the latter in a testbed setting, we report \cvpt alongside prior reinforcement learning baselines.

\textbf{Setup.} Following the Crafter protocol \citep{hafner_benchmarking_2022}, all agents get pixel inputs and are scored on Crafter Score (logarithmic average of unlocked achievements; \cref{sec:crafter_metrics}) and Return (normalized cumulative rewards), while the expert uses state inputs. \cvpt models are trained on \playdata and evaluated over 100 episodes from one training seed; Expert Policy statistics span four seeds (\cref{sec:expert_seed_diversity}). We report baseline results from their original papers.

\textbf{Results.} \cref{tbl:crafter_scores} summarizes the results. The expert policy achieves near-perfect scores, nearly doubling human performance, using a single RTX~4090 in approximately 10 hours. This demonstrates that our Expert Behavior Generator can produce high-quality demonstrations cost-efficiently, supporting large-scale behavioral pretraining in Crafter.

When trained on \playdata, \cvpt achieves strong performance among pixel-based baselines, matching the Expert Policy's Crafter Score. Training the experts with four seeds per reward instead of one does not substantially change \cvpt or \csteveone performance (\cref{sec:expert_seed_diversity}), so one seed per reward suffices.
These results establish \cvpt as a behavioral foundation model for Crafter, enabling controlled studies of planning and exploration with pre-trained priors.

Scaling from tiny to base yields a $+2.1$ gain in Crafter Score. \cref{fig:all_achievement_score} (\cref{cvpt_spec}) reports success rates on all 22 achievements.
See \cref{sec:expert_death_penalty_ablation,sec:cvpt_noop_ablation,sec:cvpt_scaling_ablation,sec:cvpt_behavior_ablation,sec:cvpt_expert_scale} for ablations.

\textbf{Qualitative Analysis of Emergent Behaviors.} Beyond quantitative metrics, \playdata captures emergent behaviors not explicitly tied to the 22 Crafter achievements, such as building shelters and bridges (\cref{sec:emergent_behavior}). 
This diversity helps \cvpt acquire general-purpose priors beyond predefined achievements, reinforcing its role as a reusable foundation for agent studies in Crafter.

\subsection{Evaluating Vision-Language Model: \cclip} \label{sec:cclip_eval}

\begin{table}[t]
\begin{minipage}[t]{0.48\linewidth}
\centering
\caption{Retrieval over the 61 training captions, queried by one held-out paraphrase per caption. Metrics include Recall at ranks 5 and 10 (R@$k$) and Mean Rank (MeanR).}
\label{tbl:cclip_eval_paraphrase}
\adjustbox{max width=\linewidth}{
\begin{tabular}{lccc}
\toprule
\textbf{Model} & \textbf{R@5} & \textbf{R@10} & \textbf{MeanR} \\
\midrule
CLIP4Clip          & $11.21\%$ & $20.25\%$ & $27.79$ \\
MineCLIP           & $16.52\%$ & $28.31\%$ & $25.64$ \\
\midrule
\cclip\,(no para.) & $79.81\%$ & $91.46\%$ & $4.11$  \\
\cclip\,(ach.)     & $53.52\%$ & $60.00\%$ & $13.43$ \\
\textbf{\cclip}    & $\mathbf{91.34\%}$ & $\mathbf{95.92\%}$ & $\mathbf{2.45}$  \\
\bottomrule
\end{tabular}
}
\end{minipage}\hfill
\begin{minipage}[t]{0.48\linewidth}
\centering
\caption{Retrieval over the 61 training captions plus one target caption for each of ten novel out-of-rule captions, queried by 100 videos per novel caption. Per-caption results in \cref{sec:cclip_unseen_per_caption}.}
\label{tbl:cclip_eval_unseen}
\adjustbox{max width=\linewidth}{
\begin{tabular}{lccc}
\toprule
\textbf{Model} & \textbf{R@5} & \textbf{R@10} & \textbf{MeanR} \\
\midrule
CLIP4Clip          & $6.40\%$  & $19.80\%$ & $27.59$ \\
MineCLIP           & $11.30\%$ & $25.50\%$ & $24.97$ \\
\midrule
\cclip\,(no para.) & $31.80\%$ & $78.80\%$ & $7.97$  \\
\cclip\,(ach.)     & $9.30\%$  & $30.60\%$ & $22.10$ \\
\textbf{\cclip}    & $\mathbf{83.60\%}$ & $\mathbf{98.60\%}$ & $\mathbf{3.88}$  \\
\bottomrule
\end{tabular}
}
\end{minipage}
\end{table}

To assess whether \ours supports vision-language grounding research, we evaluate both the quality of \clipcaption and \cclip as a grounding model via video--text retrieval.

\textbf{Setup.} We measure retrieval performance on 1{,}000 held-out episodes from \clipcaption using Recall@$k$ ($k=5,10$) and Mean Rank (MeanR) over 61 captions. 
We compare \cclip against CLIP4Clip~\citep{luo_clip4clip_2021}, a VideoCLIP trained on the general WebVid dataset~\citep{bain_frozen_2021}, and MineCLIP~\citep{fan_minedojo_2022}, a video-text model pretrained on Minecraft gameplay videos.
We additionally evaluate two \cclip ablations: \cclip\,(no para.), trained on the 61 captions without paraphrases, and \cclip\,(ach.), trained only on labels derived from the 22 Crafter achievements with paraphrase augmentation.

To test whether \cclip generalizes across paraphrases of the 61 training captions, we evaluate retrieval on held-out paraphrases of those captions, with the same 61 captions as the candidate pool. To test whether \cclip extends to captions outside the rule-generated caption set used to construct \clipcaption, we evaluate retrieval on ten novel out-of-rule captions, with the 61 training captions plus the corresponding novel caption as the candidate pool.

\textbf{Results.} As shown in \cref{tbl:cclip_eval_paraphrase}, general-purpose video-text models fall short in Crafter: CLIP4Clip and MineCLIP reach R@10 of $20.25\%$ and $28.31\%$, whereas \cclip reaches $95.92\%$, demonstrating that \cclip retrieves by caption meaning rather than by exact training wording. \cref{tbl:cclip_eval_unseen} further shows that this advantage holds on novel out-of-rule captions, indicating that the instruction vocabulary in Crafter can be extended at inference time without retraining the caption generator.

Two ablations attribute these gains to specific design choices. \cclip\,(no para.) maintains paraphrase performance but drops sharply on novel captions, isolating LLM-based paraphrase augmentation as the mechanism that carries grounding outside the rule-generated caption set. \cclip\,(ach.) collapses in both settings, indicating that achievement labels alone are insufficient and that broad coverage from the full 15 caption rules is needed.
See \cref{sec:cclip_paraphrasing_ablation,sec:cclip_unseen_per_caption} for additional ablations.

Together, these results establish \cclip as a vision-language grounding foundation for studying language-conditioned perception and instruction following in Crafter.

\subsection{Evaluating Instruction-Following Agent: \csteveone} \label{sec:csteveone_eval}

\begin{figure*}[t]
    \centering
    \includegraphics[width=1.0\textwidth]{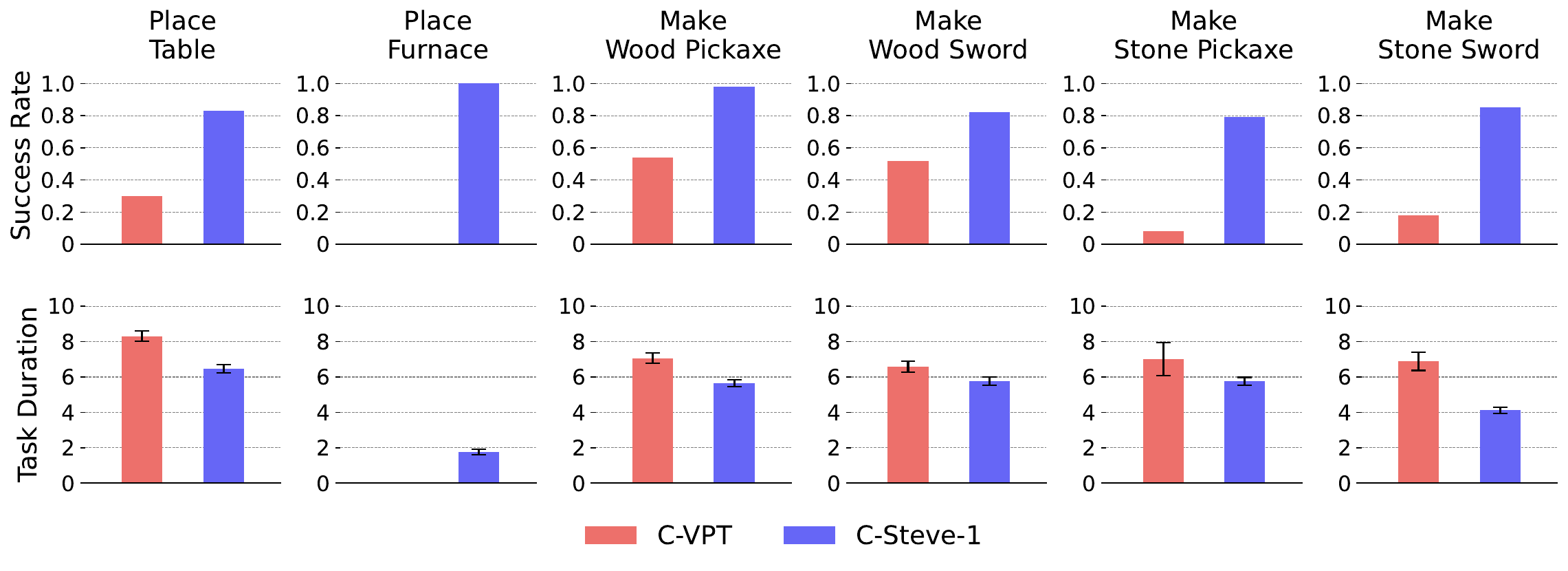}
    \caption{\csteveone Evaluation with Single Instruction on six tasks under a tight 10-step per-task budget, reporting success rates and mean and standard error of task completion time. \csteveone achieves near-perfect success rates with short task completion time. In contrast, \cvpt, which is an unconditional agent and cannot be directed to a specific task within the tight budget, shows poor performance and longer completion time, highlighting the instruction-following ability of \csteveone.}
    \label{fig:c-steve1_single_eval}
\end{figure*}

We evaluate whether CrafterDojo enables instruction-following agent research in Crafter by testing \csteveone as a language-conditioned policy.
We compare it against the unconditional \cvpt to verify that task execution is driven by language instructions rather than by behavioral priors alone.

\textbf{Setup.} We evaluate six short-horizon tasks (place crafting table, place furnace, make wood pickaxe, make wood sword, make stone pickaxe, make stone sword) with prerequisite items pre-equipped and a tight 10-step budget per task.
For instance, \emph{make stone sword} pre-equips three wood and one stone, so an instruction-follower can craft within the budget, while off-task behavior easily exhausts the 10 steps. Pre-equipping removes the confound with prerequisite acquisition, and the short budget forces goal-directed behavior rather than unconditional prior replay. This setting isolates instruction following rather than open-ended behavior, whereas \cref{sec:hierarchical_agent_eval} starts agents from an empty inventory.
We run 100 episodes per task and measure success rate and completion time in environment steps.

\textbf{Results.} \cref{fig:c-steve1_single_eval} reports the performance of \csteveone on the six tasks, compared with \cvpt as a baseline.
The gap is widest on \emph{place furnace} and the stone-tool crafts, where the unconditional prior rarely reaches the target action by chance.
\csteveone consistently outperforms the unconditional \cvpt with shorter completion times, showing that task behavior is driven by language instructions rather than behavioral priors alone.
See \cref{sec:csteveone_cfg_ablation,sec:csteveone_goal_conditioning_ablation,sec:csteveone_event_relabeling_ablation,sec:csteveone_surprise_weighting_ablation,sec:csteveone_vpt_init_ablation} for ablations.

Together with \cvpt and \cclip, these results establish the core components required for instruction-conditioned, foundation-model-based agent research in Crafter. 
We next demonstrate that these foundation models can be composed into foundation-model-based agents in this environment.

\subsection{Enabling Foundation Model-Based Agents} \label{sec:hierarchical_agent_eval}

\begin{table}[t]
    \centering
    \caption{Foundation Model-Based Agent Evaluation. Agents start with an empty inventory and must acquire all prerequisites of the target item. Success rates over 100 episodes. Flat agents receive only the final goal, without intermediate subgoals. Hierarchical agents pair a high-level planner with a low-level executor: \dep plans with an LLM, whereas \hdep plans heuristically. Composing \csteveone with a high-level planner enables long-horizon success that flat agents rarely achieve.}
    \label{tbl:long-horizon-planning-tasks}
    \small
    \begin{tabular}{lccc}
        \toprule
        \textbf{Agent} & \textbf{Iron Pickaxe} & \textbf{Iron Sword} & \textbf{Diamond} \\
        \midrule
        \multicolumn{4}{l}{\textit{Flat}} \\
        \midrule
        \csteveone & 0\% & 0\% & 1\% \\
        SPRING & 0\% & 0\% & 0\% \\
        \midrule
        \multicolumn{4}{l}{\textit{Hierarchical}} \\
        \midrule
        \dep + \csteveone & 22\% & 41\% & \textbf{7\%} \\
        \hdep + SPRING & 5\% & 7\% & 0\% \\
        \hdep + \csteveone & \textbf{44\%} & \textbf{46\%} & \textbf{7\%} \\
        \bottomrule
    \end{tabular}
\end{table}

To evaluate whether \ours enables foundation-model-based agent research in Crafter, we adopt a representative paradigm from Minecraft: hierarchical agents~\citep{wang_describe_2023} that combine an instruction-following policy with a high-level planner.
This evaluation focuses on compositionality: long-horizon Crafter goals require prerequisite acquisition that is difficult to induce from a single final-target instruction, so we test whether \csteveone can be composed with high-level planners for such tasks.

\textbf{Setup.} We select three long-horizon tasks with multiple prerequisites---Iron Pickaxe, Iron Sword, and Diamond---where agents start with an empty inventory and must acquire all prerequisites to reach the final goal. The time limits are 300 steps for Iron Pickaxe and Iron Sword and 500 steps for Diamond, and a trial succeeds if the agent obtains the target item within the time limit. We compare flat and hierarchical agents. The flat agents are \csteveone conditioned directly on the final-target instruction and SPRING~\citep{wu_spring_2023}, an LLM agent that selects primitive actions directly using Qwen3.5-27B. The hierarchical agents pair a high-level planner with \csteveone as the low-level executor: \textbf{\dep} (\cref{sec:dep_details}) follows DEP~\citep{wang_describe_2023} with Qwen3.5-27B as the planner (other open-weight LLMs in \cref{sec:llm_planner_comparison}), and \textbf{\hdep} (\cref{sec:hdep_details}) uses a heuristic planner. To test whether \csteveone is necessary as the executor, we also pair \hdep with SPRING. Each agent runs 100 episodes per task.

\textbf{Results.} \cref{tbl:long-horizon-planning-tasks} reports success rates. Both flat agents achieve near-zero success, while the hierarchical agents with \csteveone outperform them. Across planners, \hdep leads on Iron Pickaxe and Iron Sword and matches \dep on Diamond; \dep sometimes produces Minecraft-style plans that misuse Crafter's fixed tables and furnaces. Holding the \hdep planner fixed, replacing SPRING with \csteveone as the executor raises success from 5\% to 44\% on Iron Pickaxe, from 7\% to 46\% on Iron Sword, and from 0\% to 7\% on Diamond, showing that the learned execution of \csteveone is essential. The gap from \cref{sec:csteveone_eval} is expected: \csteveone is trained with goals relabeled within short event segments (\cref{sec:csteveone}), so it learns to reach the next event rather than to chain multiple prerequisites from a single final-target instruction (\cref{sec:task_difficulty_comp}); the high-level planner supplies this chain.

Together, these results show that \ours's foundation models can be composed into functional hierarchical agents---a representative paradigm in Minecraft foundation-model-based research---and that \ours supports such studies in a controlled and lightweight environment.

\section{Discussion} \label{sec:discussion}
\textbf{Limitations.} \ours focuses on Crafter as a lightweight and controllable setting for rapid iteration and systematic analysis. This focus abstracts away key sources of difficulty in larger environments such as Minecraft, including 3D visual perception, continuous camera control, and substantially longer execution horizons (\cref{sec:why_crafter}). Crafter nevertheless retains partial observability, procedural generation, sparse rewards, and long prerequisite chains, so memory, exploration, planning, and hierarchical control remain substantive challenges. Results obtained with \ours therefore provide evidence on structurally similar problems, whereas their direct transfer to Minecraft-scale environments remains a separate empirical question.

As \playdata and \clipcaption are generated automatically, their coverage depends on the expert policies, reward designs, and caption rules in the toolkits. The Diamond results also show that long-horizon composition remains challenging. Future work can extend the toolkits with additional expert objectives, broader behaviors, richer caption rules, and new downstream evaluation tasks.

\textbf{Broader Impacts.} \ours is a research infrastructure contribution for studying foundation-model-based embodied agents in simulated environments. Its expected positive impact is to lower the cost of controlled experimentation on long-horizon planning, instruction following, and vision-language grounding. The work does not involve human subjects or personal data, and the released datasets are synthetically generated in Craftax-Classic. A possible negative impact is indirect: methods developed with this stack could later be adapted to real-world or safety-critical embodied systems, such as robotics or autonomous decision making. Such extensions require careful evaluation, transparent benchmarking, and appropriate safety review before deployment outside simulation.

\section{Conclusion}
In this paper, we introduced \ours, a suite of foundation models, datasets, and toolkits that establishes Crafter as a practical testbed for foundation-model-based embodied agent research, mirroring the research ecosystem of Minecraft.
We provide \playdata for expert demonstrations, \clipcaption for vision-language grounding, and three foundation models: \cvpt for behavioral priors, \cclip for vision-language grounding, and \csteveone for instruction-following.
Our experiments demonstrate that these toolkits and datasets enable scalable data generation and effective foundation model training, with each model reliably serving its intended role.
\ours thus provides a controlled, lightweight platform for systematic studies of embodied intelligence that enables rapid iteration before scaling to complex environments like Minecraft.

\begin{ack}

This work was supported by Basic Research Laboratory Program (No. RS-2024-00414822) through the National Research Foundation of Korea (NRF) grant funded by the Korea government (MSIT). 
This work was also supported by Institute of Information \& communications Technology Planning \& Evaluation (IITP) grants funded by the Korea government (MSIT) (No. RS-2024-00509279, Global AI Frontier Lab; No. RS-2026-25617720, AI Star Fellowship (KAIST)). 
This work was also supported by Electronics and Telecommunications Research Institute (ETRI) grant funded by the Korean government [26ZR1100, Research on Intelligent Industrial Convergence].
We would like to thank the members of Machine Learning and Mind Lab (MLML) and Taewook Kang for helpful discussions and assistance.

\end{ack}

\newpage
\bibliography{refs_jyp, refs_jyp_local}
\bibliographystyle{plainnat}

%%%%%%%%%%%%%%%%%%%%%%%%%%%%%%%%%%%%%%%%%%%%%%%%%%%%%%%%%%%%

\appendix
\crefalias{section}{appendix}
\crefalias{subsection}{appendix}
\crefalias{subsubsection}{appendix}

\newpage

\section{Usage of Large Language Models}

During the preparation of this manuscript, the authors utilized large language models to refine the writing, improve grammar, and enhance overall readability. The authors critically reviewed, verified, and edited all content, and assume full responsibility for the final version of the manuscript.
Beyond writing assistance, our method-level LLM use is limited to \texttt{o4-mini-high} for \clipcaption paraphrasing (\cref{sec:llm_clip_paraphrasing}) and open-weight LLMs as the \dep planner (\cref{sec:dep_details}).

\section{Crafter Environment Specification} \label{sec:crafter_spec}

\begin{figure}[th]
\centering
\includegraphics[width=0.4\linewidth]{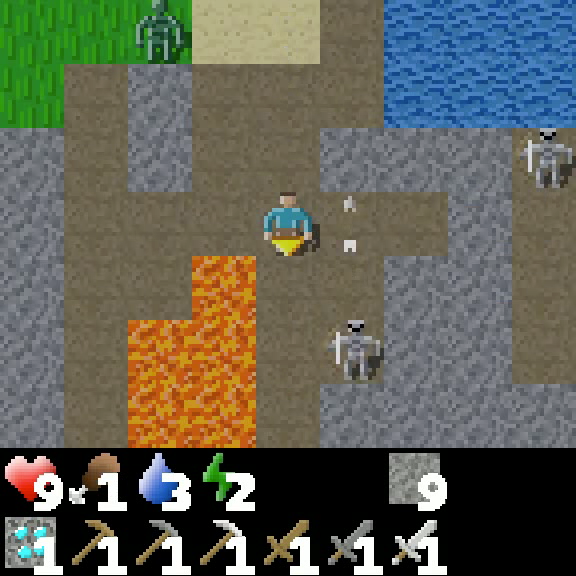}
\caption{Crafter Observation Example. Crafter is a grid-based 2D top-down environment. In Crafter, the agents \crafterplayer need to explore, craft, and collect, while surviving from hostile mobs (zombies \crafterzombie and skeletons \crafterskeleton) and maintaining life-related factors: hunger \crafterfood, thirst \crafterdrink, and energy \crafterenergy.}
\label{fig:crafter_example}
\end{figure}

Crafter~\citep{hafner_benchmarking_2022} is a 2D top-down grid-based environment with a default map size of $64 \times 64$ tiles with procedural map generation and 22 achievement tasks. These tasks require exploration, survival, and long-horizon planning to complete successfully. As a fully open-source Python environment, Crafter enables open-ended, flexible extensibility, allowing researchers to easily customize map and introduce new environment mechanics, items, and tasks.

\subsection{Observation Space}

In our experiments, we used pixel-based observations from Craftax-Classic, which uses a rendering mechanism closely matched to that of the original Crafter. \cref{fig:crafter_example} shows an example observation image. Each observation image is rendered as a $144 \times 144$ pixel frame that presents the agent and its surrounding $9 \times 7$ tiles, with an inventory panel positioned at the bottom showing the agent's current status and the types and quantities of items in the agent's inventory.

\subsection{Action Space}

Crafter features a discrete action space consisting of 17 actions. These include a \textbf{no-operation} action, four \textbf{directional movement} actions (up, down, left, right), a \textbf{do} action for interacting with other entities, a \textbf{sleep} action, four \textbf{place} actions, and six \textbf{craft} actions for creating pickaxes and swords.

Do action is treated as a no-operation if the agent is not facing an interactable entity. Place and craft actions are also treated as a no-operation if the agent has insufficient required items in the inventory. This action space design is similar to that of MineDojo~\citep{fan_minedojo_2022}, excluding camera manipulation actions. Like MineDojo, Crafter treats crafting and placing as functional actions.

\subsection{Crafter Evaluation Metrics} \label{sec:crafter_metrics}

The Crafter evaluation protocol~\citep{hafner_benchmarking_2022} covers 22 achievements spanning resource collection, crafting, survival, combat, and construction, and measures performance with Crafter Score and Return.

Crafter Score is the geometric mean of success rates over the 22 achievements, scaled to $[0, 100]$:
\begin{equation}
    S = \exp \left( \frac{1}{22} \sum_{i=1}^{22} \ln (1 + s_i) \right) - 1,
\end{equation}
where $s_i$ represents the success rate of each achievement $i$.

Return is the cumulative sum of rewards. The Crafter environment provides agents $+1$ reward when it unlocks a new achievement for the first time during an episode, $-0.1$ reward for every health point lost, and $+0.1$ reward for every health point gained. Given that the maximum return is 22, normalized return values are expressed as percentages where $100\%$ corresponds to maximum reward sum of 22.

\newpage
\section{Why Crafter? Comparison to Minecraft for Foundation Model-Based Agent Research} \label{sec:why_crafter}

Crafter is designed to support controllable and efficient research on foundation model-based agents, prioritizing research productivity over visual or environmental complexity. While Minecraft~\citep{guss_minerl_2019, fan_minedojo_2022} provides a realistic and diverse 3D environment with rich dynamics, its complexity can increase the cost of systematic analysis and slow down iteration cycles. Moreover, in Minecraft-based research environments, access to the full environment state is limited and modifying the environment often requires substantial engineering effort. This makes it challenging to perform rapid ablation, detailed failure analysis, and large-scale experiment during early-stage research.

In contrast, Crafter offers a lightweight yet expressive 2D grid-based environment whose mechanics are deliberately simplified, designed for research productivity~\citep{hafner_benchmarking_2022}. By reducing low-level perceptual and action-execution overhead while retaining challenges of long-horizon planning, Crafter enables rapid prototyping and ablation-friendly experimentation. Implemented entirely in pure Python, the environment is easy to inspect, modify, and extend, facilitating the creation of environment variants and task extensions~\citep{xu_active_2023, tang_mars_2024, stanic_learning_2024} and providing access to full environment state for systematic episode analysis and fine-grained failure diagnosis. In this sense, Crafter serves a role analogous to MNIST in vision: a controlled benchmark for sanity checks and rapid iteration before scaling up.

We compare Crafter and Minecraft along two aspects: (1) Environment Speed (\cref{sec:environment_speed_comp}), and (2) Planning Complexity and Execution Cost (\cref{sec:task_difficulty_comp}).

\subsection{Environment Speed Comparison} \label{sec:environment_speed_comp}

\begin{figure}[h]
\centering
\includegraphics[width=\linewidth]{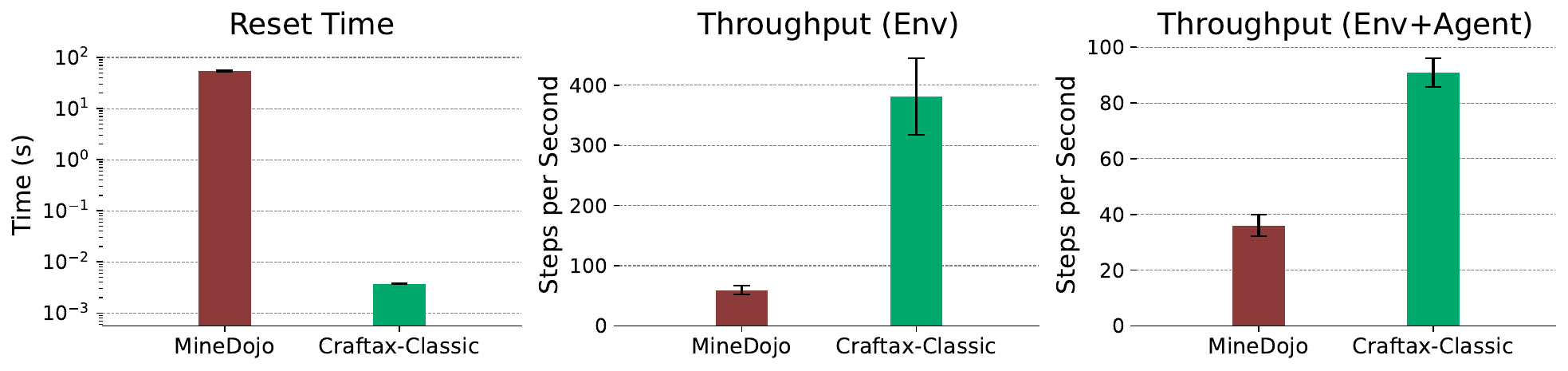}
\caption{Environment speed comparison between MineDojo and Craftax-Classic. We compare reset time and throughput of the environment alone (Env) and with the agent (Env+Agent), where the agent is VPT for MineDojo and \cvpt for Crafter. Error bars: standard deviation. Craftax-Classic shows substantially faster environment reset and higher throughput than MineDojo.}
\label{fig:speed_comparison}
\end{figure}

We compare the speed of Crafter and Minecraft (MineDojo) on the same machine with Intel(R) Xeon(R) Gold 6348 CPU and a single NVIDIA RTX 4090 GPU. We evaluate two foundation models, VPT~\citep{baker_video_2022} and \cvpt, corresponding to Minecraft and Crafter, respectively. For Minecraft, we use the smallest available model (\texttt{foundation-model-1x}, 71M parameters), while for Crafter we use the base model (15.9M parameters). All timings are measured over 10 episodes.

\cref{fig:speed_comparison} shows that, compared to MineDojo, Crafter exhibits lower reset time and higher throughput in both the environment-only (Env) setting and the setting including agent inference (Env+Agent). The higher environment-only throughput indicates that the Crafter simulator itself is more lightweight and computationally efficient than MineDojo, which is built on top of the full Minecraft game engine, and this advantage persists when including agent inference despite Crafter using a substantially smaller foundation model. Together, these properties make Crafter a computationally efficient platform for rapid iteration on foundation model-based agents.

\subsection{Planning Complexity and Execution Cost Comparison} \label{sec:task_difficulty_comp}

\begin{table}[h]
\centering
\caption{Planning complexity and execution cost comparison between Minecraft and Crafter for obtaining major items. Prerequisites and Actions are the number of distinct item types required and the median number of environment steps. Minecraft quantities are from~\citet{baker_video_2022}, and those for Crafter are based on our expert policy. We additionally report the estimated wall-clock time required for VPT and \cvpt agents to obtain each item, computed from the speed measurements in \cref{fig:speed_comparison}.}
\begin{tabular}{@{}ccccccc@{}}
\toprule
\multirow{2}{*}{\textbf{Item}} & \multicolumn{3}{c}{\textbf{Minecraft}}                & \multicolumn{3}{c}{\textbf{Craftax-Classic}} \\ \cmidrule(lr){2-4}\cmidrule(lr){5-7} 
                  & \textbf{Prerequisites} & \textbf{Actions} & \multicolumn{1}{c}{\textbf{Time (s)}} & \textbf{Prerequisites}   & \textbf{Actions} & \textbf{Time (s)}  \\ \midrule
Wood              & 0             & 280    & \multicolumn{1}{c}{7.8}    & 0               & 9    & 0.1     \\
Stone             & 5             & 2,050  & \multicolumn{1}{c}{56.9}   & 3               & 32   & 0.4     \\
Iron              & 7             & 6,540  & \multicolumn{1}{c}{181.5}  & 5               & 111  & 1.2     \\
Diamond           & 11            & 23,975 & \multicolumn{1}{c}{665.4}  & 9               & 712  & 7.8    \\ \bottomrule
\end{tabular}
\label{tbl:task_difficulty_comp}
\end{table}

\cref{tbl:task_difficulty_comp} shows a comparison of planning complexity and execution cost between Minecraft and Crafter for the four item obtaining tasks. In Minecraft, the tasks are obtaining log \minecraftlog, cobblestone \minecraftstone, iron ore \minecraftsironore, and diamond \minecraftdiamond, while in Crafter the corresponding tasks are obtaining wood \crafterwood, stone \crafterstone, iron \crafteriron, and diamond \crafterdiamond. For each task, we report Prerequisites, the number of distinct item types required, and Actions, the median number of environment steps needed. We also report estimated wall-clock time, computed using speed measurements from \cref{fig:speed_comparison}.

Crafter and Minecraft differ in wall-clock time even at the easiest milestone. For wood, the estimated time is about 0.1 seconds in Crafter and 7.8 seconds in Minecraft, excluding environment reset time. This gap persists for stone, iron, and diamond. For diamond, the estimated time is 7.8 seconds in Crafter and approximately 11 minutes in Minecraft.

The comparable prerequisite counts show that Crafter preserves the task structure: agents must still plan similar subgoal sequences. The difference is execution---Crafter removes 3D perception and continuous control overhead while keeping the prerequisite structure that makes planning hard.

Overall, these results suggest that Crafter enables fast evaluation of tasks with planning complexity comparable to Minecraft. This makes Crafter well suited for early-stage research prototyping, where the goal is to iterate quickly on hypotheses, model variants, and experimental settings before scaling to more complex and realistic environments like Minecraft.

\newpage
\section{\playdata Dataset Specification} \label{sec:playdata_spec}

\begin{table}[h]
\centering
\caption{Hyperparameters for expert policy training.}
\begin{tabular}{lc}
\toprule
\textbf{Hyperparameter} & \textbf{Value}     \\
\midrule
Learning Rate          & $2 \times 10^{-4}$ \\
Discount Factor        & $0.99$             \\
Num Steps              & $64$               \\
Batch Size             & $8$                \\
Epochs                 & $4$                \\
Max Grad Norm          & $1.0$              \\
Num Environments       & $4096$             \\
Total Timesteps        & 10B                \\
\bottomrule
\end{tabular}
\label{tbl:expert_policy_hyperparameters}
\end{table}

\subsection{Generation Process}

We used the Expert Behavior Generator toolkit to generate \playdata from PPO-RNN expert policies trained in Craftax-Classic-Symbolic~\citep{matthews_craftax_2024}. For the released dataset, we trained two experts with different rewards, \emph{Survival} and \emph{Gathering}, and mixed their trajectories in equal proportion. This gives \playdata both long-horizon survival behavior and inventory-acquisition behavior. The generation process consists of two steps: (1) expert policy training and (2) trajectory generation.

\textbf{Expert Policy Training.} We train two PPO-RNN experts, \emph{Survival} and \emph{Gathering}, in the Craftax-Classic symbolic environment. The two experts share the same training protocol and hyperparameters in \cref{tbl:expert_policy_hyperparameters}; they differ only in the reward shaping used during training.

\textit{Shared setup.} Both experts start from the standard Craftax reward, which combines achievement progress and health. We add a death penalty of $\lambda_\text{death} = 10$ whenever the episode ends in death. This prevents the policy from terminating early after collecting the available achievement rewards.

\textit{Survival expert.} This expert uses only the shared reward above, so it learns to unlock achievements while staying alive for long episodes, yielding trajectories with long-horizon survival.

\textit{Gathering expert.} The Gathering expert uses the same survival-oriented reward, but additionally rewards inventory acquisition. Specifically, it receives a shaping term proportional to the per-step net gain over the 12 Craftax inventory slots: wood, stone, coal, iron, diamond, sapling, and the three tiers each of pickaxe and sword. Writing $I_{t,j}$ for the count of item $j$ at step $t$, the Gathering reward is
\begin{equation}
r_t = r^{\text{base}}_t + \alpha \sum_{j=1}^{12} \max\bigl(I_{t,j} - I_{t-1,j},\, 0\bigr) - \lambda_\text{death}\,\mathbf{1}\{\text{death}\}_t,
\label{eq:gatherer_reward}
\end{equation}
where $\alpha = 0.1$. Placement actions that consume an item without producing a new one (e.g., \textsc{place\_stone}, \textsc{place\_table}) are penalized symmetrically by $-\alpha$ per lost unit, so that place-then-mine cycles cannot be exploited. Crafting actions are not penalized because converting raw materials into tools is the intended behavior. The Survival expert corresponds to $\alpha=0$.

\begin{table}[tp!]
\centering
\caption{Data Structure of Each Trajectory in \playdata. $L$ denotes the episode length. $N_\text{entity}$ denotes the maximum number of entities. For cows, zombies, and arrows, the maximum number of entities is 4 each. The maximum number of skeletons is 2. The maximum number of plants is 10.}
\begin{tabular}{ll}
\toprule
\textbf{Name}                          & \textbf{Shape}                \\
\midrule                    
RGB Image                              & $(L+1, 3, 144, 144)$          \\
Action                                 & $(L,)$                        \\
\midrule                   
Map                                    & $(L+1, 64, 64)$               \\
Light Level                            & $(L+1,)$                      \\
Growing Plants Position                & $(L+1, N_\text{plants}, 2)$   \\ 
Growing Plants Status                  & $(L+1, N_\text{plants}, 2)$   \\ 
\midrule                    
Player Position                        & $(L+1, 2)$                    \\
Player Direction                       & $(L+1,)$                      \\
Player (Health/Food/Drink/Energy)      & $(L+1,4)$                     \\
Player (Recover/Hunger/Thirst/Fatigue) & $(L+1,4)$                     \\
Player Sleeping                        & $(L+1,)$                      \\
Player Inventory                       & $(L+1,12)$                    \\
Player Achievements                    & $(L+1, 22)$                   \\
\midrule            
Mob Map                                & $(L+1, 64, 64)$               \\
Cow Position                           & $(L+1, N_\text{cow}, 2)$      \\
Cow Status                             & $(L+1, N_\text{cow}, 3)$      \\
Zombie Position                        & $(L+1, N_\text{zombie}, 2)$   \\
Zombie Status                          & $(L+1, N_\text{zombie}, 3)$   \\
Skeleton Position                      & $(L+1, N_\text{skeleton}, 2)$ \\
Skeleton Status                        & $(L+1, N_\text{skeleton}, 3)$ \\
Arrow Position                         & $(L+1, N_\text{arrow}, 2)$    \\
Arrow Status                           & $(L+1, N_\text{arrow}, 3)$    \\
Arrow Direction                        & $(L+1, N_\text{arrow}, 2)$    \\
\bottomrule
\end{tabular}
\label{tbl:playdata_structure}
\end{table}

\textbf{Trajectory Generation.} Following expert policy training, we generated \playdata as 20{,}000 expert trajectories sampled in equal proportion from the Survival and Gathering experts (10{,}000 each). Each trajectory comprises three components: (1) Environment States, (2) Rendered Images corresponding to the states, and (3) Agent Action Sequence. \cref{tbl:playdata_structure} details the trajectory structure.

\subsection{Dataset Statistics}

The default \playdata comprises 20{,}000 expert trajectories (10{,}000 each from the Survival and Gathering experts) with a total of approximately 158M timesteps and an average length of 7{,}887 steps, created using the Expert Behavior Generator toolkit. At the dataset level, \playdata achieves a Crafter Score of 97.82, and 13{,}800 / 20{,}000 trajectories (69.0\%) reach the 10{,}000-step time limit. Per-achievement success rates of the Survival expert are shown in \cref{fig:all_achievement_score}.

\subsection{Emergent Behaviors of Expert Policy} \label{sec:emergent_behavior}

\begin{figure}[hbp!]
\centering
\includegraphics[width=\linewidth]{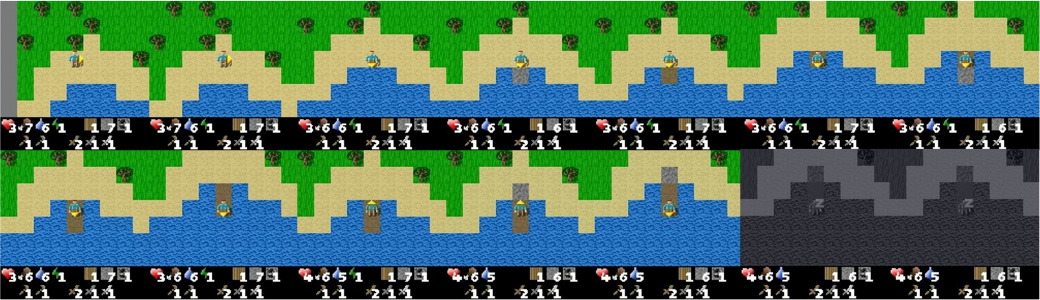}
\caption{Emergent Survival Behavior of the Expert Policy Example. When energy is low, the agent constructs a shelter on water to safely sleep while avoiding hostile mobs that spawn frequently at night and avoid increased damage to sleeping agents.}
\label{fig:emergent_behavior1}
\end{figure}

\begin{figure}[htp!]
\centering
\includegraphics[width=0.9\linewidth]{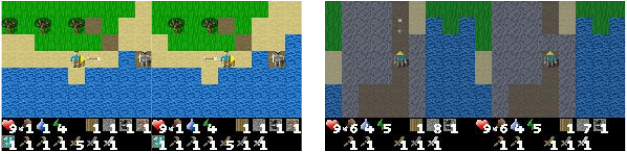}
\caption{Emergent Survival Behavior of the Expert Policy Example. \textbf{(Left)} The agent jumps strategically to avoid incoming arrows or \textbf{(Right)} blocks them by placing a stone.}
\label{fig:emergent_behavior2}
\end{figure}

\begin{figure}[htp!]
\centering
\includegraphics[width=0.8\linewidth]{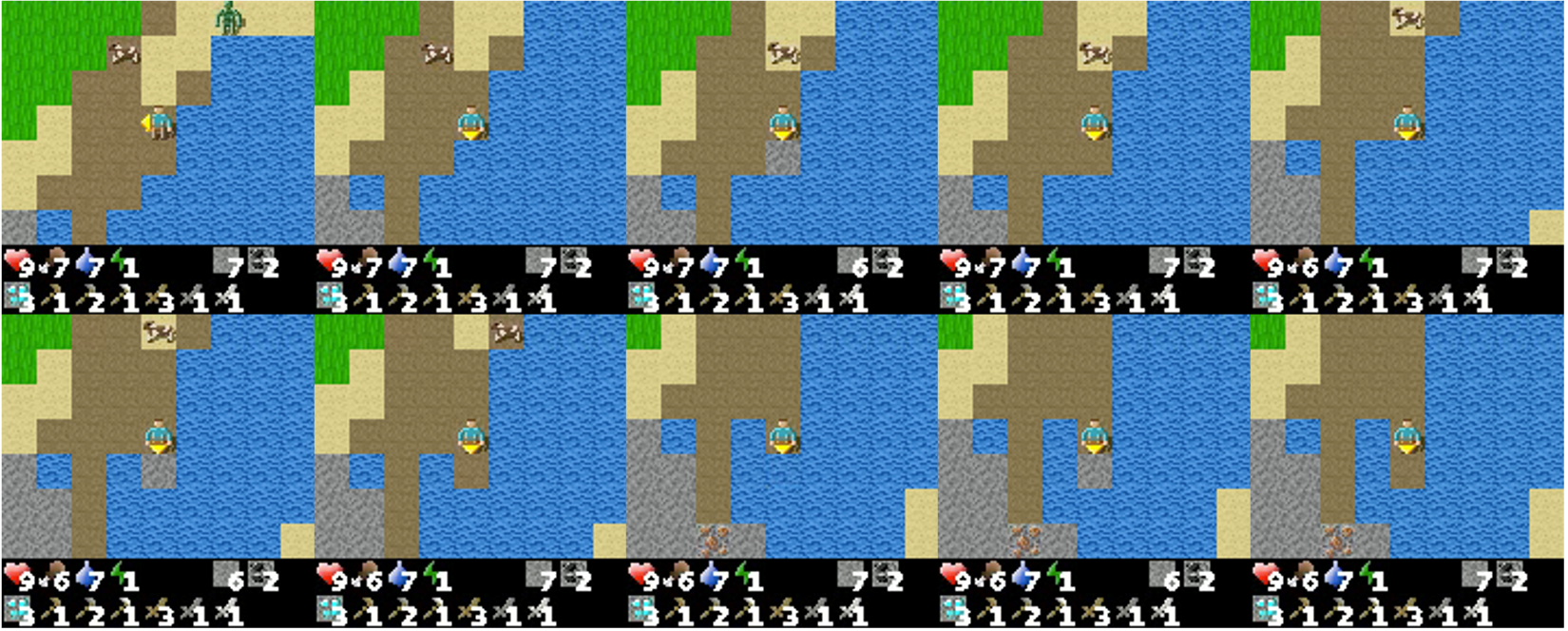}
\caption{Emergent Expert Behavior. The agent extends the ground by placing and removing stones.}
\label{fig:emergent_behavior3}
\end{figure}

The expert policy exhibits emergent behaviors that extend beyond the Crafter 22 achievements. The agent develops strategic behaviors such as constructing shelters (\cref{fig:emergent_behavior1}), strategic moving (\cref{fig:emergent_behavior2} (Left)), blocking incoming arrow (\cref{fig:emergent_behavior2} (Right)), and building road on water (\cref{fig:emergent_behavior3}).

\subsection{Expert Policy Training Ablation: Death Penalty} \label{sec:expert_death_penalty_ablation}

\begin{figure}[h]
\centering
\includegraphics[width=0.7\linewidth]{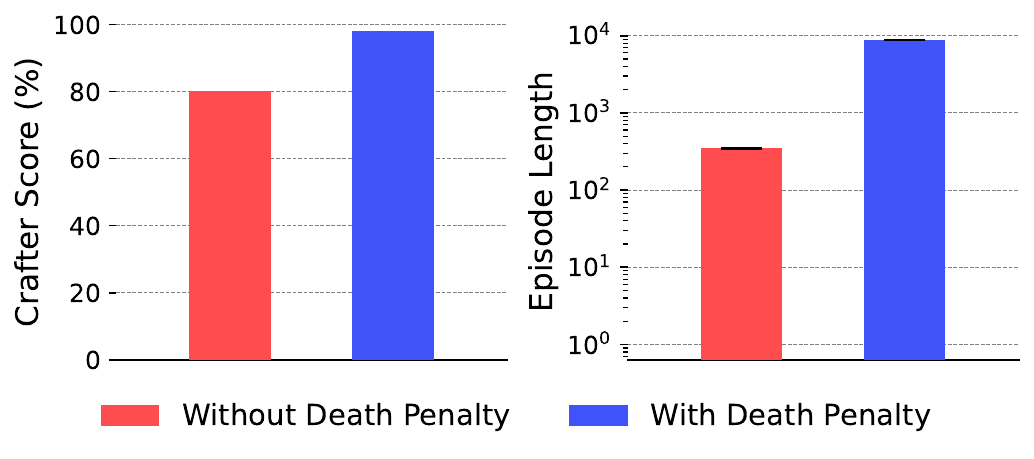}
\caption{Expert policy performance with and without the death penalty. \textbf{(Left)} Crafter Score. \textbf{(Right)} Episode Length. The death penalty improves both scores and survival time. Episode length error bars indicate standard error of the mean across rollout episodes.}
\label{fig:expert_ablation}
\end{figure}

We analyze how the death penalty affects demonstration quality, a key factor for establishing a reliable behavioral foundation in Crafter.
\cref{fig:expert_ablation} shows that adding a death penalty substantially improves both Crafter Score and episode length.
Without a death penalty, the expert terminates early once no further achievements can be unlocked.
With a coefficient of 10, the expert survives longer and scores higher, producing better demonstrations for \playdata.
Longer survival lets the expert pursue achievements with delayed rewards, such as finding diamonds or waiting for crops to ripen.

\subsection{Computational Cost}

Training each expert policy took about 10 hours on a single NVIDIA RTX 4090 GPU. Generating the 20,000 trajectories took about 5 hours on 8 RTX Pro 6000 GPUs.

\newpage
\section{\clipcaption Dataset Specification} \label{sec:clipcaption_spec}

\subsection{Generation Process}

After the \playdata was built using the Expert Behavior Generator toolkit, the Caption Generator toolkit utilizes environment states and actions to generate captions. The Caption Generator toolkit is a rule-based system with a total of 15 rules, where each rule has three attributes: name, rule checker, and caption template. \cref{tbl:caption_templates} lists each rule with its caption template and an example caption.

The caption generation system evaluates rule checkers by examining both the action sequences and the environment state, listed in \cref{tbl:playdata_structure}. For each timestep $t$, the system checks whether the combination of action and resulting environment state changes matches any of the predefined rule checkers. When a rule condition is satisfied at timestep $t$, the system adds a video sequence $o_{t-4:t+1}$ along with its associated caption to the \clipcaption dataset.

\subsection{Dataset Statistics}

\begin{figure*}[!ht]
\centering
\includegraphics[width=\textwidth]{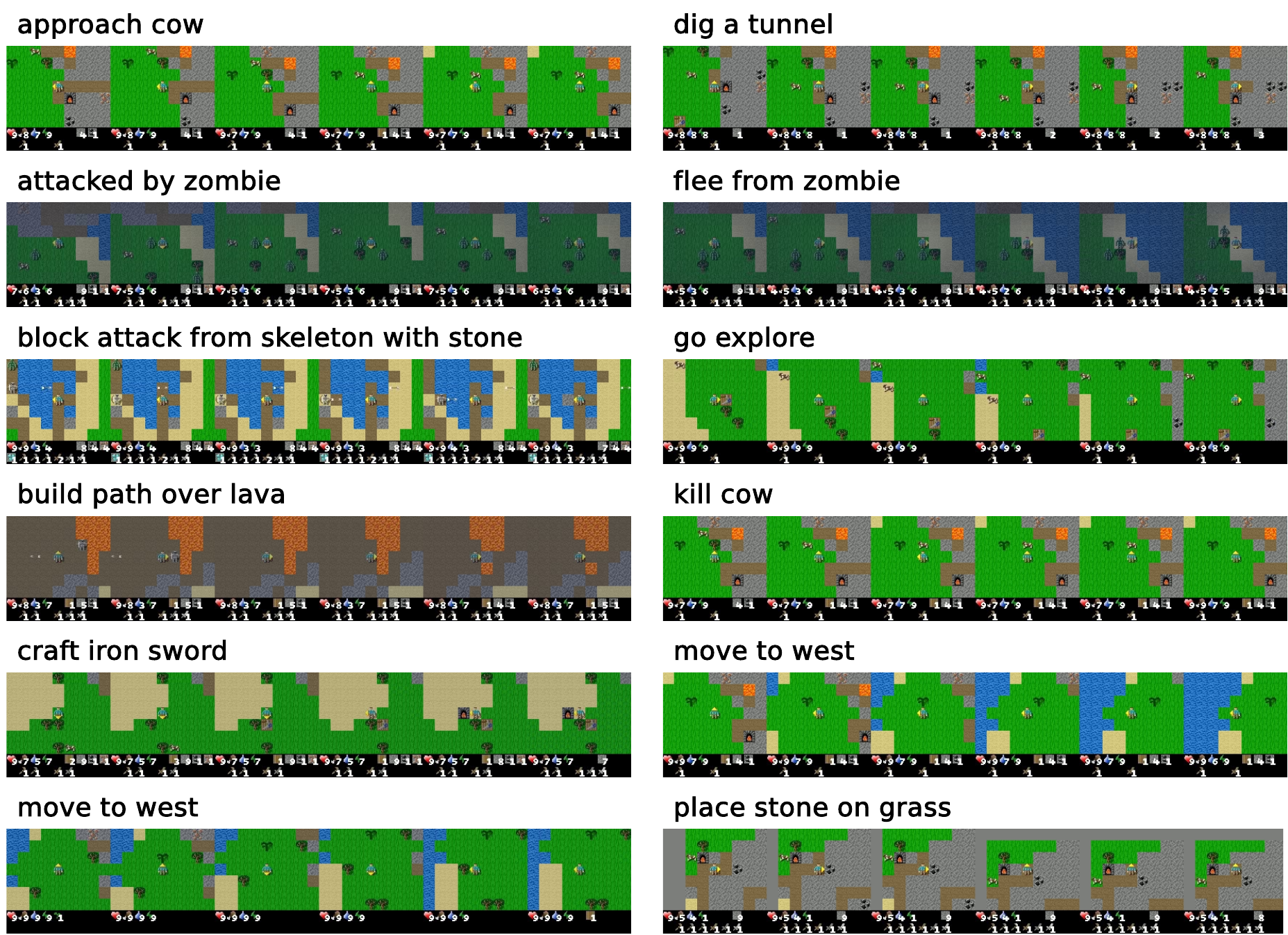}
\caption{Example of Video Segment and Caption pairs in the \clipcaption dataset.}
\label{fig:caption_samples}
\end{figure*}

Using the Caption Generator toolkit on the full \playdata, we generate approximately 67M raw video-caption pairs. Since the raw distribution is heavily skewed toward common rules, we subsample to a uniform distribution over the 15 caption rules, yielding the released \clipcaption with about 1.2M pairs. We assign each pair to the split of its source episode in \playdata: 1.1M pairs are used for training, 64K for validation, and 65K for testing. \cref{fig:caption_samples} shows example video-caption pairs.

\begin{table*}[htp!]
\centering
\caption{The 15 rules used to generate \clipcaption. The generator fills each caption template by substituting placeholders in curly braces (\textit{e.g.}, $\{\text{item}\}$, $\{\text{mob}\}$) with specific entities.}
\adjustbox{width=\textwidth}{
\begin{tabular}{@{}llll@{}}
\toprule
\textbf{Category}             & \textbf{Rule} & \textbf{Caption Template}                                & \textbf{Example Caption}            \\ \midrule
\multirow{5}{*}{Achievement}  & Harvest       & obtain $\{\text{item}\}$                                  & obtain stone                        \\
                              & Place         & place $\{\text{item}\}$ on $\{\text{material}\}$          & place table on grass                \\
                              & Craft         & craft $\{\text{item}\}$                                   & craft wood pickaxe                  \\
                              & Kill          & kill $\{\text{mob}\}$                                     & kill cow                            \\
                              & Sleep         & go to sleep                                               & go to sleep                         \\ \cmidrule(r){1-4}
\multirow{5}{*}{Movement}     & Stay          & stay                                                      & stay                                \\
                              & Move          & move to $\{\text{dir}\}$                                  & move to north                       \\
                              & Approach      & approach $\{\text{mob}\}$                                 & approach zombie                     \\
                              & Flee          & flee from $\{\text{mob}\}$                                & flee from skeleton                  \\
                              & Explore       & go explore                                                & go explore                          \\ \cmidrule(r){1-4}
\multirow{3}{*}{Construction} & Shelter       & place $\{\text{item}\}$ to build shelter                  & place stone to build shelter        \\
                              & Path          & build path over $\{\text{material}\}$                     & build path over water               \\
                              & Tunnel        & dig a tunnel                                              & dig a tunnel                        \\ \cmidrule(r){1-4}
\multirow{2}{*}{Combat}       & Attacked By   & attacked by $\{\text{mob}\}$                              & attacked by zombie                  \\
                              & Block Attack  & block attack from $\{\text{mob}\}$ with $\{\text{item}\}$ & block attack from zombie with stone \\ \bottomrule
\end{tabular}
}
\label{tbl:caption_templates}
\end{table*}

\subsection{Computational Cost}

Caption generation was performed on CPUs only, using a server equipped with two Intel Xeon Gold 6542Y CPUs (48 physical cores and 96 logical threads), and took approximately 4.6 hours.

\clearpage

\section{CrafterVPT Details} \label{cvpt_spec}

\subsection{\cvpt Training Details}

\begin{table}[h]
\begin{minipage}[t]{0.48\linewidth}
\centering
\caption{Model Hyperparameters for C-VPT.}
\label{tbl:cvpt_architectures}
\adjustbox{max width=\linewidth}{
\begin{tabular}{lcc}
\toprule
\textbf{Hyperparameter}     & \textbf{Tiny}      & \textbf{Base}         \\
\midrule
Total Parameters            & 3.2M               & 15.9M                 \\
$d_\text{model}$            & 128                & 512                   \\
$d_\text{ff}$               & 512                & 2048                  \\
Attention Heads             & 4                  & 4                     \\
Recurrence Layers           & 2                  & 4                     \\
IMPALA Width                & 1                  & 1                     \\
IMPALA Channels             & [16, 32, 32]       & [16, 32, 32]          \\
\bottomrule
\end{tabular}
}
\end{minipage}\hfill
\begin{minipage}[t]{0.48\linewidth}
\centering
\caption{Hyperparameters for Training \cvpt.}
\label{tbl:cvpt_hyperparameters}
\adjustbox{max width=\linewidth}{
\begin{tabular}{lcc}
\toprule
\textbf{Hyperparameter}     & \textbf{Tiny}      & \textbf{Base}      \\
\midrule
Optimizer                   & AdamW              & AdamW              \\
Batch Size                  & $128$              & $128$              \\
Epochs                      & $10$               & $10$               \\
Learning Rate (LR)          & $5 \times 10^{-4}$ & $5 \times 10^{-4}$ \\
LR Warmup Frames            & 10M                & 10M                \\
Weight Decay                & $1 \times 10^{-4}$ & $1 \times 10^{-4}$ \\
Max Grad Norm               & $5$                & $5$                \\
\bottomrule
\end{tabular}
}
\end{minipage}
\end{table}

We present the model architecture and training hyperparameters in \cref{tbl:cvpt_architectures,tbl:cvpt_hyperparameters}, respectively. We applied a cosine decay with linear warmup learning rate scheduler.

\subsection{Noop Action Filtering Ablation} \label{sec:cvpt_noop_ablation}

\begin{table}[h!]
\centering
\caption{No-op action filtering ablation. Both variants share the same source \playdata and identical \cvpt (base) training hyperparameters; only the no-op filtering strategy differs. Crafter Score and Return are reported as mean over 100 evaluation episodes (Return: mean $\pm$ SEM).}
\label{tab:cvpt_noop_ablation}
\begin{tabular}{lcc}
    \toprule
    \textbf{Strategy} & \textbf{Crafter Score} & \textbf{Return} \\
    \midrule
    No filtering             & $95.48$          & $0.938 \pm 0.006$ \\
    Selective filtering (ours) & $\mathbf{97.68}$ & $\mathbf{0.955 \pm 0.006}$ \\
    \bottomrule
\end{tabular}
\end{table}

\begin{figure}[h!]
\centering
\includegraphics[width=\textwidth]{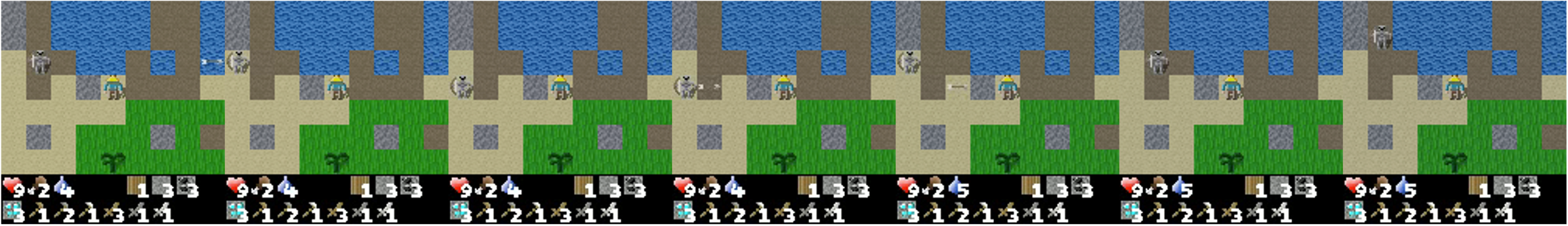}
\caption{Strategic Waiting. The agent waits beside a stone to block a skeleton's arrows.}
\label{fig:strategic_waiting}
\end{figure}

No-op actions constitute 50\% of all recorded actions, with 46.4\% of these corresponding to sleep periods (23.2\% of all timesteps overall). We compare two filtering strategies on the same \playdata, training \cvpt (base) under identical hyperparameters: \emph{No filtering} and \emph{Selective filtering} (our default, which removes no-op sequences shorter than $20$ consecutive steps and actions issued while sleeping regardless of sequence length, while otherwise preserving longer ones associated with strategic waiting behaviors such as those illustrated in \cref{fig:strategic_waiting}). \cref{tab:cvpt_noop_ablation} reports the result.

The unfiltered baseline retains too much inactive supervision and underperforms our default by $-2.20$ Crafter Score. Our selective filtering reduces the no-op ratio from $\sim$50\% to $\sim$3\% while preserving longer non-sleep no-op sequences associated with strategic waiting, yielding the better Crafter Score.

\subsection{Crafter 22 Achievements Performance Comparison}

\begin{figure}[h]
\centering
\includegraphics[width=\textwidth]{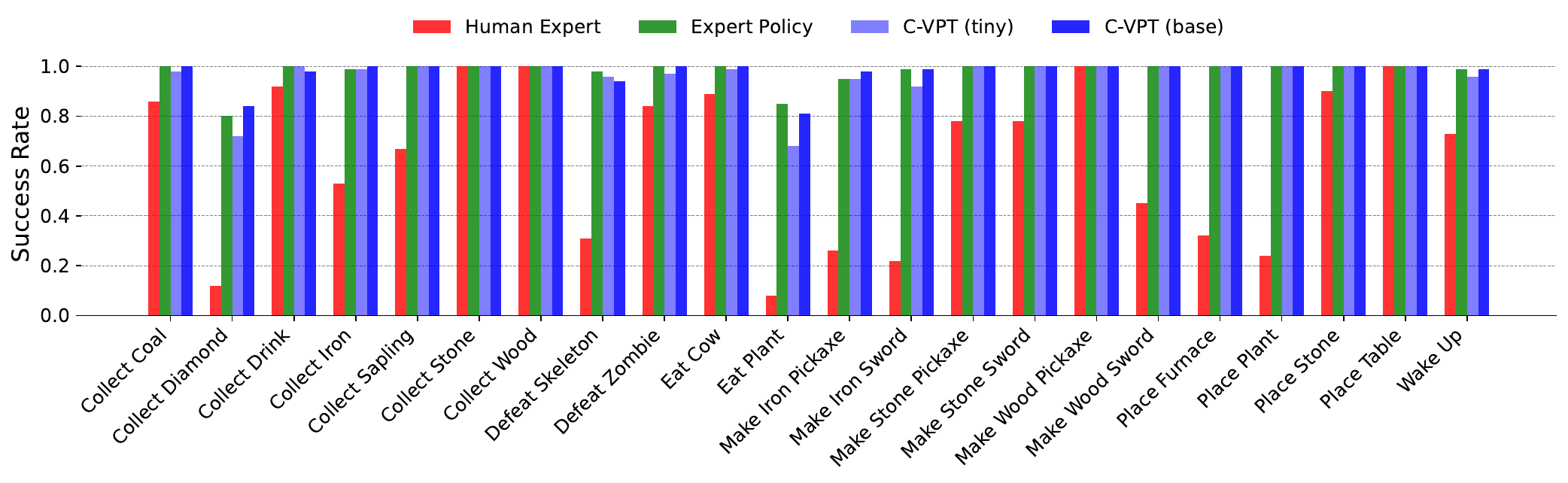}
\caption{Success Rates of Crafter 22 Achievements Comparison between Human Expert, Our learned Expert Policy, and \cvpt (tiny and base).}
\label{fig:all_achievement_score}
\end{figure}

\cref{fig:all_achievement_score} presents the success rates for all 22 Crafter achievements for the human expert, the expert policy, and \cvpt (tiny and base). The expert policy consistently outperforms both the human expert and all other models across all achievements. Meanwhile, \cvpt achieves performance that is largely comparable to the human expert, with lower success rates observed only for eat plant and collect sapling. Overall, the results suggest that \cvpt can serve as an effective behavior prior, exhibiting largely human-comparable performance across a wide range of Crafter achievements, and that the expert policy enables the scalable collection of trajectories encompassing diverse fundamental tasks.

\subsection{\playdata Scaling Ablation} \label{sec:cvpt_scaling_ablation}

\begin{figure}[h]
\centering
\includegraphics[width=0.7\linewidth]{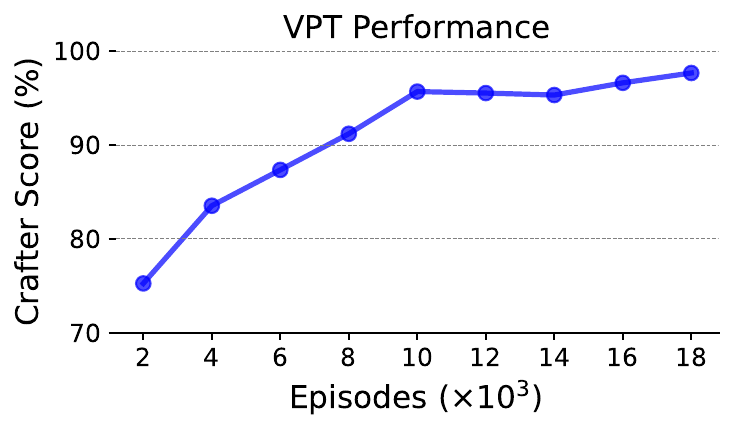}
\caption{\cvpt performance according to the size of the training dataset. Performance improves with more episodes and stabilizes around 18,000, indicating sufficient data for training \cvpt.}
\label{fig:vpt_scaling}
\end{figure}

We examine how dataset size influences the effectiveness of behavioral pretraining to determine a practical scale for a Crafter foundation dataset.
We train \cvpt (base) on subsets of \playdata with varying episode counts and evaluate performance.
In \cref{fig:vpt_scaling}, Crafter Score rises sharply until about 10{,}000--12{,}000 episodes, then largely saturates, with the full 18{,}000-episode set adding little.

% NEW (NeurIPS 2026): \cvpt ablation over \playdata behavioral composition.
\subsection{\playdata Behavioral Composition Ablation} \label{sec:cvpt_behavior_ablation}

A behavioral foundation model should provide a reusable behavioral prior, not merely imitate the highest-scoring single policy. We therefore train experts under different rewards and combine their rollouts at the dataset level. To test how this composition affects \cvpt, we compare four frame-controlled \playdata variants with comparable numbers of environment frames, and each variant is used to train a separate \cvpt (base):
\begin{itemize}[leftmargin=*,noitemsep,topsep=0pt]
    \item \textbf{Crafter}: trajectories from an expert trained on the standard Crafter reward only. This is the unshaped natural-reward baseline.
    \item \textbf{Survival}: trajectories from an expert trained on the Crafter reward augmented with the $-10$ death penalty. As shown in \cref{fig:expert_ablation}, the death penalty pushes the expert to prolong survival once all 22 achievements have been unlocked, yielding longer survival than the \textit{Crafter} reward.
    \item \textbf{Gathering}: trajectories from a Gathering expert that adds an inventory-acquisition shaping term to the Survival reward (\cref{eq:gatherer_reward}).
    \item \textbf{Mixed}: Survival and Gathering trajectories mixed with comparable frame contributions from each expert.
\end{itemize}

\begin{table}[h]
\centering
\caption{\cvpt (base) behavior under different \playdata compositions. All variants use comparable frame counts and identical training hyperparameters; only the data composition differs. Episode Length is the mean and standard error over 100 episodes. Tool Re-crafts is the mean per-episode count, summed across the six craftable tools, of crafting actions executed after each tool's first unlock.}
\label{tab:cvpt_behavior_ablation}
\begin{tabular}{lccc}
    \toprule
    \textbf{Composition} & \textbf{Crafter Score} & \textbf{Episode Length} & \textbf{Tool Re-crafts} \\
    \midrule
    Crafter    & 77.68 & $323.9 \pm 24.2$   & 0.2  \\
    Survival   & 97.81 & $7190.7 \pm 400.5$ & 7.6  \\
    Gathering  & 94.16 & $5397.6 \pm 438.9$ & 39.5 \\
    Mixed      & 97.68 & $6320.4 \pm 430.5$ & 20.0 \\
    \bottomrule
\end{tabular}
\end{table}

\cref{tab:cvpt_behavior_ablation} shows that reward shaping changes the behavioral prior inherited by \cvpt. The \textit{Crafter} expert terminates early once achievements are exhausted, so \cvpt also inherits short episodes and poor Crafter Score. The death penalty removes this early-termination bias, while the Gathering reward shifts behavior toward repeated tool crafting and item gathering. Since \textit{Mixed} keeps the long-horizon behavior of \textit{Survival} while adding part of \textit{Gathering}'s crafting tendency, we use it as the default \playdata composition for subsequent \cvpt experiments.

\subsection{Expert Policy Training Scale Ablation} \label{sec:cvpt_expert_scale}

We additionally study how the expertise level of the data-generating policy affects \cvpt. Fixing the reward to the Survival reward, we take three Survival checkpoints (Survival-500M, Survival-1B, and Survival-10B) and use each to generate a separate \playdata variant for \cvpt (base), with comparable numbers of frames in the resulting training datasets.

\begin{table}[h]
\centering
\caption{\cvpt (base) performance according to expert policy training scale. We report Crafter Score and mean and standard error of Episode Length over 100 episodes.}
\label{tab:cvpt_expert_scale}
\begin{tabular}{lcc}
    \toprule
    \textbf{Expert Checkpoint} & \textbf{Crafter Score} & \textbf{Episode Length} \\
    \midrule
    Survival-500M & 64.77 & $435.9 \pm 2.1$    \\
    Survival-1B   & 92.92 & $5527.4 \pm 320.4$ \\
    Survival-10B  & 97.81 & $7190.7 \pm 400.5$ \\
    \bottomrule
\end{tabular}
\end{table}

\cref{tab:cvpt_expert_scale} isolates expert maturity by keeping training-set frame counts comparable. The earliest Survival checkpoint transfers an incomplete prior: before the expert reliably reaches late-game achievements, \cvpt inherits short-horizon behavior and weak achievement coverage. More mature checkpoints provide stronger long-horizon routines, with the converged expert giving the best behavioral foundation. Thus, even at a fixed data scale, source-expert quality still matters.

\subsection{Computational Cost}

We employed 4 NVIDIA RTX 4090 GPUs for \cvpt (tiny) and \cvpt (base) variants. The training time was 7 hours for \cvpt (tiny) and 8.5 hours for \cvpt (base).

\newpage
\section{CrafterCLIP Details} \label{sec:cclip_spec}

\begin{table}[h]
\centering
\caption{Hyperparameters for \cclip Training.}
\begin{tabular}{ll}
\toprule
\textbf{Hyperparameter}          & \textbf{Value}      \\
\midrule
Optimizer                        & AdamW               \\
Batch Size                       & 128                 \\
Epoch                            & $2$                 \\
Learning Rate                    & $1.5 \times 10^{-4}$ \\
LR Warmup Steps                  & 100                 \\
Weight Decay                     & 0.2                 \\
Layerwise LR Decay               &  $0.65$             \\
Pre-trained Layers LR Multiplier &  $0.5$              \\
\bottomrule
\end{tabular}
\label{tbl:mineclip_hyperparameters}
\end{table}

\subsection{Training Details}

Since the rule frequencies in the original \clipcaption are heavily skewed, we subsampled to a uniform distribution over the 15 caption rules. We therefore trained \cclip on about 1.1M video-caption pairs from the 18{,}000-episode training split. The full \clipcaption specification, including the raw-to-balanced subsampling and the per-episode split assignment, is in \cref{sec:clipcaption_spec}.

We employ a cosine decay with linear warmup learning rate scheduler for training \cclip with the hyperparameters in \cref{tbl:mineclip_hyperparameters}. We also apply a \textbf{layer-wise learning rate decay} across the layers of both the vision and text encoders, following the training of MineCLIP~\citep{fan_minedojo_2022}.

\subsection{LLM-based Caption Rephrasing} \label{sec:llm_clip_paraphrasing}

To enhance the linguistic diversity, we generate paraphrased captions by prompting \texttt{o4-mini-high} with the prompt in \cref{lst:caption_rephrase_prompt}, inspired by LaCLIP \cite{fan_improving_2023}. The paraphrased captions are stored as a dictionary, mapping original template-based captions to lists of paraphrased caption variants. During \cclip training, we use the original caption as a key to retrieve its paraphrased list and uniformly random sample one caption variant for each iteration.

\vspace{1em}
\begin{lstlisting}[caption=Caption Rephrasing Prompt, label=lst:caption_rephrase_prompt]
<goal>
rewrite each video caption. generate 50 unique captions for each caption.
</goal>
<guidelines>
- focus on the behavior, do not introduce a new tools like sword if there is no sword in the original captions.
- avoid generating title-like sentences.
</guidelines>
<format>
"original 1":
  - "rephrased 1-1"
  - "rephrased 1-2"
"original 2":
  - "rephrased 2-1"
  - "rephrased 2-2"
</format>
<captions>
- approach cow
...
</captions>
\end{lstlisting}

\textbf{Examples.} We provide paraphrased caption samples in YAML format in \cref{lst:paraphrased_caption_samples}. These samples are generated following the format guidelines specified in \cref{lst:caption_rephrase_prompt}, where each original template-based caption is mapped to multiple paraphrased caption variants.

\vspace{1em}
\begin{lstlisting}[caption=Paraphrased Caption Samples, label=lst:paraphrased_caption_samples]
"attacked by skeleton":
  - "agent is struck by the skeleton"
  - "take a hit from the skeleton"
  - "agent gets hit by the skeleton"
"block attack from skeleton with crafting table":
  - "agent blocks the skeleton's attack with a crafting table"
  - "use crafting table to block skeleton attack"
  - "hold up crafting table against skeleton strike"
"craft iron sword":
  - "agent crafts an iron sword"
  - "forging an iron sword"
  - "agent creates an iron sword"
"flee from zombie":
  - "agent flees from a zombie"
  - "fleeing a zombie"
  - "agent runs from a zombie"
\end{lstlisting}

\subsection{LLM-based Paraphrasing Ablation} \label{sec:cclip_paraphrasing_ablation}

\begin{figure}[h]
    \centering
    \includegraphics[width=0.6\linewidth]{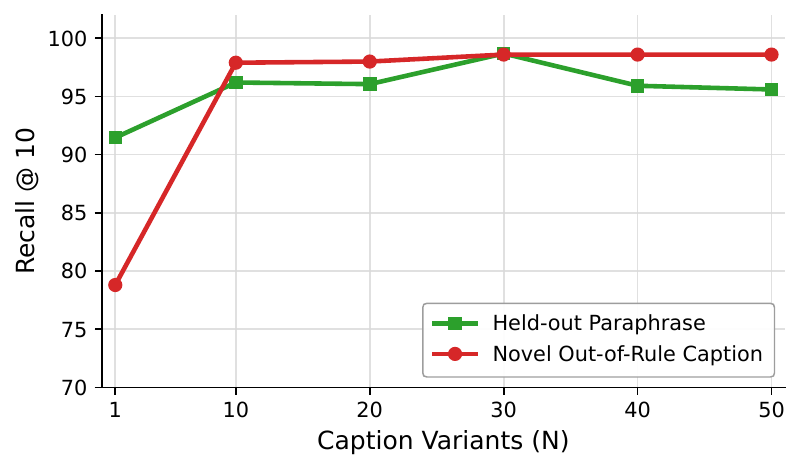}
    \caption{Recall@10 of \cclip\ as a function of the number of caption variants per template $N$ ($1$ original $+$ $(N{-}1)$ LLM paraphrases), in the LLM-paraphrasing scaling ablation. The held-out paraphrase curve uses the split of \cref{tbl:cclip_eval_paraphrase}; the out-of-rule curve uses the ten novel out-of-rule captions.}
    \label{fig:crafterclip_scaling}
\end{figure}

We study how LLM-paraphrased augmentation in \clipcaption\ affects \cclip\ retrieval by retraining \cclip\ with different numbers of caption variants per template and evaluating both held-out paraphrases and the novel out-of-rule captions in \cref{tbl:cclip_eval_unseen}. The trend is simple: held-out paraphrases saturate with only modest augmentation, whereas novel out-of-rule captions benefit from broader lexical variation before reaching a plateau. Since the out-of-rule split better reflects inference-time instruction variation, we set $N=40$ as the default caption-variant count.

\subsection{Per-Caption Details on the Novel Out-of-Rule Caption Split} \label{sec:cclip_unseen_per_caption}

\begin{table}[h]
\centering
\caption{Per-caption retrieval results on the ten novel out-of-rule captions summarized in \cref{tbl:cclip_eval_unseen}, with the \cclip ablations from \cref{tbl:cclip_eval_paraphrase,tbl:cclip_eval_unseen} broken down by novel out-of-rule caption. Each novel out-of-rule caption contributes 100 frame queries, retrieved against the 61 training captions plus itself. Values are R@5 / R@10 (\%, higher is better) and MeanR (lower is better).}
\label{tbl:cclip_unseen_per_caption}
\adjustbox{max width=\linewidth}{
\begin{tabular}{lcccccccccccc}
\toprule
& \multicolumn{3}{c}{\textbf{CLIP4Clip}} & \multicolumn{3}{c}{\textbf{\cclip\,(no para.)}} & \multicolumn{3}{c}{\textbf{\cclip\,(ach.)}} & \multicolumn{3}{c}{\textbf{\cclip}} \\ \cmidrule(lr){2-4}\cmidrule(lr){5-7}\cmidrule(lr){8-10}\cmidrule(lr){11-13}
\textbf{Novel out-of-rule caption} & R@5 & R@10 & MeanR & R@5 & R@10 & MeanR & R@5 & R@10 & MeanR & R@5 & R@10 & MeanR \\
\midrule
find ocean          & $26.0$ & $63.0$ & $9.91$  & $61.0$ & $93.0$  & $5.48$  & $9.0$  & $38.0$ & $25.72$ & $95.0$ & $100.0$ & $2.44$ \\
find desert         & $31.0$ & $70.0$ & $8.69$  & $20.0$ & $61.0$  & $10.11$ & $4.0$  & $47.0$ & $17.16$ & $90.0$ & $99.0$  & $3.44$ \\
find forest         & $0.0$  & $4.0$  & $23.49$ & $68.0$ & $100.0$ & $5.17$  & $78.0$ & $99.0$ & $5.08$  & $69.0$ & $99.0$  & $4.97$ \\
find grassland      & $0.0$  & $0.0$  & $28.85$ & $69.0$ & $99.0$  & $5.24$  & $1.0$  & $58.0$ & $13.69$ & $75.0$ & $99.0$  & $4.52$ \\
barren landscape    & $0.0$  & $1.0$  & $37.07$ & $14.0$ & $60.0$  & $9.86$  & $0.0$  & $2.0$  & $30.12$ & $91.0$ & $99.0$  & $3.44$ \\
edge of desert      & $0.0$  & $1.0$  & $39.65$ & $11.0$ & $50.0$  & $11.83$ & $0.0$  & $3.0$  & $32.16$ & $70.0$ & $96.0$  & $4.90$ \\
peaceful meadow     & $0.0$  & $0.0$  & $55.87$ & $39.0$ & $94.0$  & $6.58$  & $0.0$  & $38.0$ & $14.77$ & $72.0$ & $94.0$  & $5.50$ \\
low health          & $5.0$  & $34.0$ & $12.38$ & $6.0$  & $88.0$  & $7.71$  & $1.0$  & $10.0$ & $23.43$ & $88.0$ & $100.0$ & $3.91$ \\
dangerous situation & $2.0$  & $25.0$ & $15.08$ & $29.0$ & $77.0$  & $7.79$  & $0.0$  & $11.0$ & $21.51$ & $90.0$ & $100.0$ & $2.88$ \\
safe resting spot   & $0.0$  & $0.0$  & $44.88$ & $1.0$  & $66.0$  & $9.93$  & $0.0$  & $0.0$  & $37.37$ & $96.0$ & $100.0$ & $2.82$ \\
\bottomrule
\end{tabular}
}
\end{table}

\subsection{Computational Cost}

We trained \cclip\ on a single NVIDIA RTX Pro 6000 GPU for 1.5 hours.

\newpage
\section{CrafterSteve-1 Details} \label{sec:csteveone_spec}

\subsection{Training Details}
\begin{table}[h]
\centering
\caption{Hyperparameters for training \csteveone.}
\begin{tabular}{ll}
\toprule
\textbf{Hyperparameters}                  & \textbf{Value}     \\
\midrule
Batch Size                                & 128                \\
Total Training Frames                     & $2 \times 10^8$    \\
Sequence Length                           & 640                \\
Truncated Length                          & 64                 \\
Learning Rate                             & $1 \times 10^{-5}$ \\
LR Warmup Frames                          & $1 \times 10^7$    \\
Weight Decay                              & $10^{-4}$          \\
Max Grad Norm                             & 5                  \\
\midrule
Min / Max Steps Between Goals             & 1 / 10             \\
Uncond. Goal Sampling Probability         & $0.1$              \\
NLL Clip Threshold ($c_\text{clamp}$)     & $2.83$             \\
\bottomrule
\end{tabular}
\label{tbl:hyperparameter_steveone}
\end{table}

\cref{tbl:hyperparameter_steveone} lists the \csteveone training hyperparameters. Our code follows the original Steve-1 implementation~\citep{lifshitz_steve-1_2023}, except event-based packed hindsight relabeling and goal conditioning.

\subsection{Classifier-Free Guidance Scale Ablation} \label{sec:csteveone_cfg_ablation}
\begin{figure}[h]
    \centering
    \includegraphics[width=\textwidth]{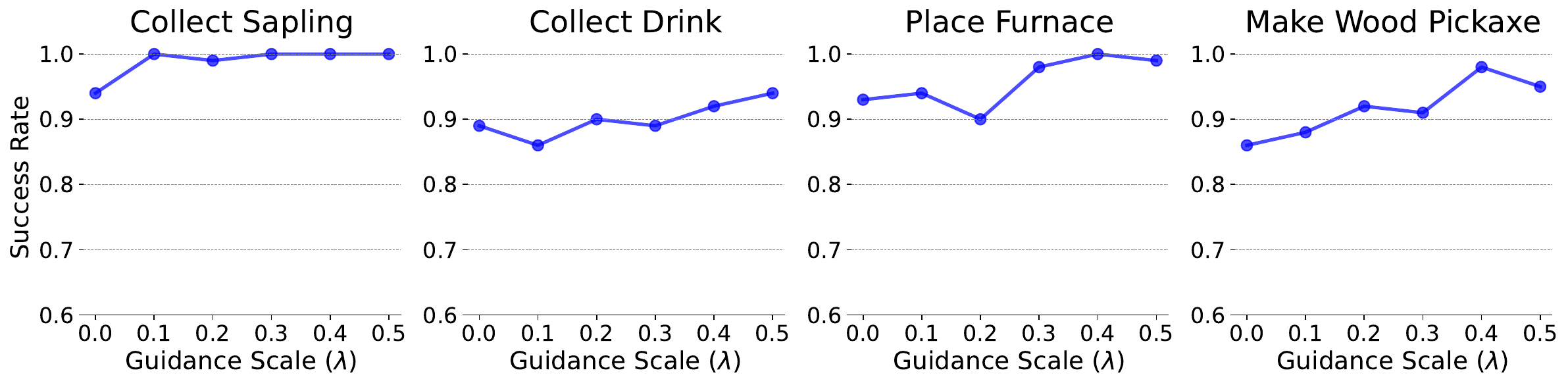}
    \caption{\csteveone\ success rates versus the classifier-free guidance scale $\lambda$ on four tasks: \textit{Collect Sapling}, \textit{Collect Drink}, \textit{Place Furnace}, and \textit{Make Wood Pickaxe}, averaged over 100 episodes.}
\label{fig:cfg_ablation}
\end{figure}

\cref{fig:cfg_ablation} sweeps the classifier-free guidance scale across four representative tasks. In this limited ablation, adding a moderate amount of conditioning guidance improves over the unconditional policy, particularly for tasks that require a specific placement or crafting action. Performance is stable near the middle of the sweep, and we use $\lambda=0.4$ by default in the main experiments.

\subsection{Goal-Conditioning Ablation} \label{sec:csteveone_goal_conditioning_ablation}
We evaluate two goal conditioning methods for instruction-following agents: TransformerXL Conditioning (\textbf{TrXL Cond}) from the original Steve-1~\cite{lifshitz_steve-1_2023}, and Policy Head Conditioning (\textbf{Head Cond}), introduced in MrSteve~\cite{park_mrsteve_2025}. \cref{tbl:c-steve1_conditioning_eval} presents success rates across six single-goal tasks. Head Cond consistently outperforms TrXL Cond in both full fine-tuning and LoRA fine-tuning.

\begin{table}[h]
\centering
\caption{Two conditioning method comparison on six short-horizon tasks (place crafting table, place furnace, make wood pickaxe, make wood sword, make stone pickaxe, make stone sword) under a tight 10-step per-task budget. TrXL Conditioning (TrXL Cond) was introduced from original Steve-1~\citep{lifshitz_steve-1_2023}, while Policy Head Conditioning (Head Cond) was introduced from MrSteve~\citep{park_mrsteve_2025}.}
\begin{tabular}{ccccc}
\toprule
\multirow{2}{*}{\textbf{Task}} & \multicolumn{2}{c}{TrXL Cond}       & \multicolumn{2}{c}{Head Cond}               \\ \cmidrule(lr){2-3}\cmidrule(lr){4-5}
                      & \multicolumn{1}{c}{Full}   & LoRA   & \multicolumn{1}{c}{Full}   & LoRA            \\ \midrule
Place Crafting Table  & \multicolumn{1}{c}{38\%}   & 69\%   & \multicolumn{1}{c}{58\%}   & 83\%            \\
Place Furnace         & \multicolumn{1}{c}{11\%}   & 74\%   & \multicolumn{1}{c}{57\%}   & 100\%           \\
Make Wood Pickaxe     & \multicolumn{1}{c}{29\%}   & 61\%   & \multicolumn{1}{c}{75\%}   & 98\%            \\
Make Wood Sword       & \multicolumn{1}{c}{58\%}   & 78\%   & \multicolumn{1}{c}{71\%}   & 82\%            \\
Make Stone Pickaxe    & \multicolumn{1}{c}{15\%}   & 34\%   & \multicolumn{1}{c}{72\%}   & 79\%            \\
Make Stone Sword      & \multicolumn{1}{c}{8\%}    & 33\%   & \multicolumn{1}{c}{69\%}   & 85\%            \\ \cmidrule{1-5}
Average               & \multicolumn{1}{c}{26.5\%} & 58.2\% & \multicolumn{1}{c}{67.0\%} & \textbf{87.8\%} \\
\bottomrule
\end{tabular}
\label{tbl:c-steve1_conditioning_eval}
\end{table}

\newpage
\subsection{Event-based Hindsight Relabeling Ablation} \label{sec:csteveone_event_relabeling_ablation}

\begin{figure}[h]
\centering
\includegraphics[width=\textwidth]{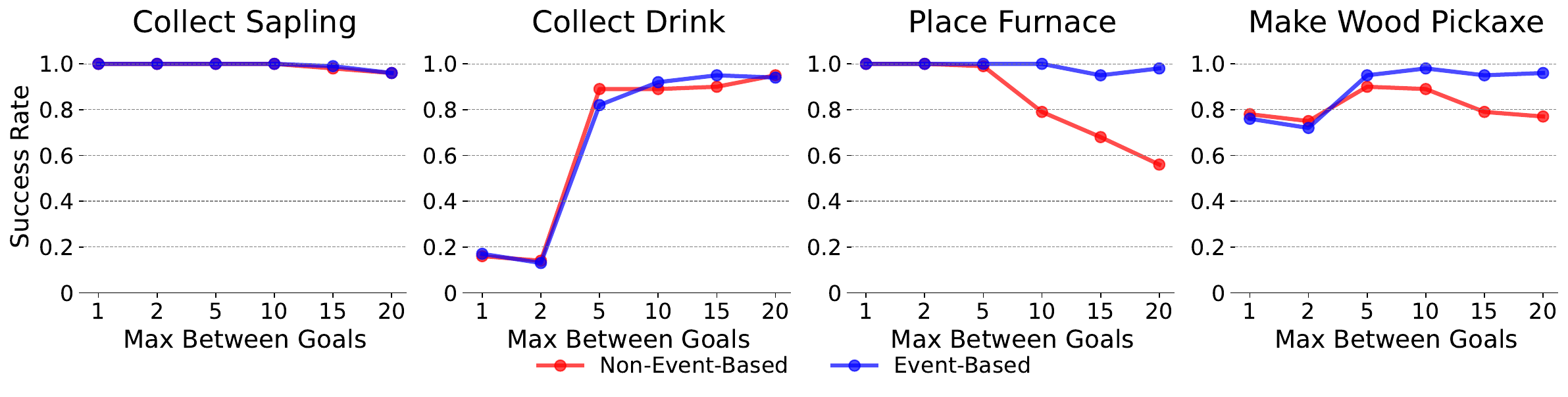}
\caption{Comparison between non-event-based and event-based packed hindsight relabeling.}
\label{fig:mxbtn_ablation}
\end{figure}

To evaluate event-based packed hindsight relabeling, we compare it with the non-event-based variant across maximum goal distances on four representative tasks (short-horizon collection, long-horizon navigation, placement, and basic crafting). In \cref{fig:mxbtn_ablation}, the event-based variant matches or exceeds the non-event-based one on every task and stays stable as the maximum goal distance grows.

The advantage is most visible on tasks whose useful goal horizons do not align with a fixed step interval. Event-based relabeling groups goals by coherent environment changes, so sampled goals stay meaningful across collection, navigation, placement, and crafting. This makes relabeling less sensitive to the maximum goal distance, and we use the stable middle setting by default.

\subsection{Surprise Weighting Ablation} \label{sec:csteveone_surprise_weighting_ablation}

\begin{table}[h]
\centering
\caption{Surprise weighting (SW) ablation: \csteveone success rates on six tasks over 100 episodes.}
\begin{tabular}{lcc}
\toprule
\textbf{Task}         & No-SW   & SW              \\ \midrule
Collect Sapling       & 100\%   & 100\%           \\
Collect Drink         & 97\%    & 92\%            \\
Place Crafting Table  & 56\%    & 83\%            \\
Defeat Zombie         & 96\%    & 95\%            \\
Make Wood Pickaxe     & 72\%    & 98\%            \\
Make Stone Pickaxe    & 83\%    & 79\%            \\ \midrule
Average               & 84.0\%  & \textbf{91.2\%} \\
\bottomrule
\end{tabular}
\label{tbl:c-steve1_sw_eval}
\end{table}

\cref{tbl:c-steve1_sw_eval} shows that Surprise Weighting mainly helps tasks where the policy must learn a specific, low-frequency action sequence, so we used SW in the default \csteveone.

\subsection{VPT Initialization Ablation} \label{sec:csteveone_vpt_init_ablation}

\begin{table}[h]
\centering
\caption{VPT initialization ablation under the \hdep planner on the same three long-horizon tasks as \cref{tbl:long-horizon-planning-tasks}, evaluated over 100 episodes per task. The four \csteveone\ variants differ only in the source VPT policy; all other training and inference settings are identical. \cvpt\ compositions follow \cref{sec:cvpt_behavior_ablation}, with \emph{Mixed} being the default composition adopted in the main paper.}
\begin{tabular}{lccc}
\toprule
\textbf{\csteveone\ initialization}    & \textbf{Iron Pickaxe} & \textbf{Iron Sword} & \textbf{Diamond} \\ \midrule
From scratch                            & 1\%                   & 0\%                 & 0\%              \\
From \cvpt\ (Gathering)                 & 11\%                  & 6\%                 & 0\%              \\
From \cvpt\ (Survival)                  & 28\%                  & 34\%                & 3\%              \\
From \cvpt\ (Mixed; default)            & \textbf{44\%}         & \textbf{46\%}       & \textbf{7\%}     \\
\bottomrule
\end{tabular}
\label{tbl:c-steve1_vpt_init_eval}
\end{table}

Our default \csteveone\ recipe initializes the LoRA-augmented TransformerXL from \cvpt\ trained on the \emph{Mixed} \playdata composition (\cref{sec:cvpt_behavior_ablation}). To isolate the role of this behavioral prior, we compare \csteveone\ variants initialized from \cvpt\ (Mixed), \cvpt\ (Survival), \cvpt\ (Gathering), or from scratch, while keeping the training setup and \hdep evaluation fixed.

\cref{tbl:c-steve1_vpt_init_eval} shows that hierarchical planning still depends on a strong low-level behavioral prior. A planner can propose the right subgoals, but executing them requires the dense action distribution learned by \cvpt; training the controller from scratch is not enough. Among the \cvpt\ sources, the Mixed prior is the most useful for long-horizon execution, consistent with its combination of Survival's long-horizon behavior and Gathering's inventory-acquisition bias.

\subsection{Computational Cost}

We used 8 NVIDIA RTX 4090 GPUs for about an hour to train \csteveone.

\newpage

\section{Reference Hierarchical Planning Agent Details} \label{sec:full_agent_details}

To demonstrate that \ours enables foundation model-based agent research in Crafter, we build reference hierarchical agents following the Minecraft paradigm~\citep{wang_describe_2023, wang_jarvis-1_2023, li_optimus-1_2024}. Each agent is composed of a high-level planner, which decomposes a long-horizon task into a sequence of subgoals, and a low-level controller, which executes each subgoal through primitive actions.

We define a subgoal as a short-horizon objective. In Crafter, subgoals fall into three categories: mining, crafting, and placing a target object. Unlike long-horizon tasks, each subgoal assumes that all required tools and items were prepared by preceding subgoals. We use \csteveone as the low-level controller, showing how the foundation models fit the hierarchical agent paradigm.

The high-level planner decides which subgoal to execute next. Concretely, the planner outputs one of three structured subgoals: \texttt{mine(item)}, \texttt{craft(item)}, or \texttt{place(item)}. The planner is invoked when there is no active subgoal, or when the previous subgoal successfully or unsuccessfully terminates, following DEPS~\citep{wang_describe_2023}. Each structured subgoal is parsed into the corresponding controller-level specification and executed by the low-level controller.

We implement two high-level planners, one LLM-based and one heuristic. The following sections describe the LLM-based high-level planner (\cref{sec:dep_details}), the heuristic high-level planner (\cref{sec:hdep_details}), and an example Diamond-task plan produced by the LLM-based planner (\cref{sec:high_level_plan_examples}).

\subsection{LLM-Based High-Level Planner Details} \label{sec:dep_details}

The system prompt (\cref{lst:dep_system_prompt}) provides domain knowledge including available items and entities, and the planning query prompt (\cref{lst:dep_planning_prompt}) specifies the current task and inventory state.

\vspace{1em}
\begin{lstlisting}[caption=DEP Planner System Prompt, label=lst:dep_system_prompt]
You are an AI assistant that helps the user complete tasks in the Crafter environment, which is a Minecraft-like crafting/mining environment, using a code-planning format.
Your job is to generate corrected and updated plans when the user reports failures.
You do not execute the steps; you only produce revised code-plans.

# Rules
## 1. Action
- You may use only the following actions in your plans:
    - mine(output_item, tools)
    - craft(output_item, input_items, tools)
    - place(block_item)
- output_item is a single item string like 'wood' or 'stone_pickaxe'. You must only produce one item per mine/craft action; repeat actions if you need multiples.
- input_items is a dictionary like {'wood': 3, 'stone': 1}.
- tools is either a list of tool names (e.g., ["stone_pickaxe"], ["table", "furnace"]) or null.

## Tool Requirements
- Mining some materials requires correct tools.
- Crafting some items requires specific tools (e.g., table, furnace)
- If the user fails a step because of missing items or missing tools, you must identify the cause and add the required preparatory steps.

## 3. Inventory Awareness
- The user will state their current inventory when a step fails.
- You must:
    - Treat the user's provided inventory as the authoritative current state.
    - Adjust your plan so that every required tool and ingredient is provided before it needed.
    - Never assume items exist that the user did not explicitly say they have.

## 4. Correction Cycle
1. Determine why the step failed (missing tool, missing ingredients, incorrect tool, etc.).
2. Insert the necessary prerequisite steps before the failed step.
3. Regenerate the entire updated plan beginning from the current inventory the user has stated.
4. Produce code in the format:
```python
def <task_name>(inventory = <current_inventory>):
    <action 1>
    <action 2>
    ...
    return '<target_item>'

- Please note that the order of your plan will be executed step-by-step. Be aware of the order of each action.

## 5. DO NOT include examples
- Only produce the updated code solution when requested.
- DO NOT include meta-explanation unless helping user understand a failure.

## 6. Available Item/Entity List
- Use the following item names for your planning: wood, stone, coal, iron, diamond, table, furnace, sapling, drink, wood_pickaxe, wood_sword, stone_pickaxe, stone_sword, iron_pickaxe, iron_sword, plant, zombie, skeleton, cow
\end{lstlisting}

\begin{lstlisting}[caption=DEP Planner Planning Prompt Example, label=lst:dep_planning_prompt]
I have items in inventory: 9 health, 9 food, 9 drink, 9 energy. You cannot see a table or furnace nearby. How to obtain diamond?
Plan as executable code:
```python
def obtain_diamond(inventory = {{}}):
    <1st action>
    <2nd action>
    <3rd action>
    ...
    return 'diamond'
```

Remember: each mine/craft action must produce exactly one output_item (count = 1). Repeat actions if you need multiple of the same item.
\end{lstlisting}

\begin{lstlisting}[caption=Crafter Knowledge Prompt, label=lst:crafter_knowledge]
# About Crafter
Crafter is a 2D open-world survival environment. Your goal is to maximize your survival time while manage your vital signs, gather resources, craft tools, and defend yourself against hostile creatures.

1. Vital Signs & Status
- Health: You start with 9/9 Health. You lose health when attacked by enemies or if your Food/Drink levels drop to zero. Health regenerates over time if your Food and Drink levels are high.
- Food: You start with 9/9 Food. It decreases over time. Replenish it by hunting cows or eating plants.
- Drink: You start with 9/9 Drink. It decreases over time. Replenish it by drinking from water tiles.
- Energy: Used for performing actions. Replenish it by sleeping.
- Inventory: You have an inventory to store gathered resources (Wood, Stone, Coal, Iron, Diamond, Saplings) and crafted items (Tools, Stations) up to 9 items each.

2. Basic Mechanics
- Grid World: The world is a 2D grid containing materials like Grass, Water, Stone, Sand, Path, Lava, and Trees.
- Movement: You can move Up, Down, Left, or Right (move_up, move_down, move_left, move_right).
- Interaction: Use the do action to interact with the object or material directly in front of you. This includes attacking, gathering, eating, or drinking.
- Sleeping: Use the sleep action to rest. This restores Energy and safely skips through the dangerous night, provided you are not in immediate danger.

3. Resource Gathering & Tool Hierarchy You need specific tools to harvest harder materials.
- Wood: Harvest Trees (Requires: None). Yields 1 Wood.
- Saplings: Harvest Grass (Requires: None). Low probability (10%) to yield 1 Sapling.
- Water: Interact with Water (Requires: None). Replenishes Drink level.
- Stone: Harvest Stone (Requires: Wood Pickaxe). Yields 1 Stone.
- Coal: Harvest Coal (Requires: Wood Pickaxe). Yields 1 Coal.
- Iron: Harvest Iron (Requires: Stone Pickaxe). Yields 1 Iron.
- Diamond: Harvest Diamond (Requires: Iron Pickaxe). Yields 1 Diamond.

4. Crafting & Stations Most advanced items require a crafting station to be placed nearby.
- Crafting Table:
    - Recipe: 2 Wood.
    - Usage: Required to craft Wood/Stone Pickaxes and Swords.
    - !IMPORTANT! Crafting Table cannot be stored in the inventory, as it is NOT an item.
- Furnace:
    - Recipe: 1 Stone.
    - Usage: Required (alongside a Table) to craft Iron Pickaxes and Swords.
    - !IMPORTANT! Furnace cannot be stored in the inventory, as it is NOT an item.
    - Tip: When furnace is required, placing both table and furnace is strongly recommended.
- Unlike Minecraft, table and furnace cannot be broken, you may need to frequently make them.

5. Tool & Weapon Recipes
- Wood Pickaxe: 1 Wood (Requires: Table).
- Stone Pickaxe: 1 Wood + 1 Stone (Requires: Table).
- Iron Pickaxe: 1 Wood + 1 Coal + 1 Iron (Requires: Table + Furnace).
- Wood Sword: 1 Wood (Requires: Table).
- Stone Sword: 1 Wood + 1 Stone (Requires: Table).
- Iron Sword: 1 Wood + 1 Coal + 1 Iron (Requires: Table + Furnace).

6. Combat & Enemies
- Zombies: Spawn on Grass, primarily at night. They chase and attack you.
- Skeletons: Spawn on Paths. They shoot arrows from a distance.
- Defense: You can attack enemies using the do action. Swords significantly increase your damage output compared to bare hands or pickaxes.
- Strategy: Nighttime is dangerous due to increased spawn rates. Building shelter or sleeping is recommended.

7. Farming & Building
- Place Plant: 1 Sapling. Grows into a mature plant over time and, after 300 steps, it can be eaten.
- Place Stone: 1 Stone. Creates a wall to block enemies.
- Cows: Passive mobs. Can be hunted for Food.

8. Achievements (Goals) Unlock these by performing the specific actions:
- Gathering: Collect Wood, Stone, Coal, Iron, Diamond, Drink, Sapling.
- Crafting: Make Wood/Stone/Iron Pickaxe, Make Wood/Stone/Iron Sword, Place Table, Place Furnace.
- Farming/Building: Place Plant, Place Stone, Eat Plant, Wake Up.
- Combat: Defeat Zombie, Defeat Skeleton.
- Survival: Eat Cow.
\end{lstlisting}

\subsection{Heuristic High-Level Planner Details} \label{sec:hdep_details}

The heuristic planner generates plans from a knowledge graph of item dependencies in Crafter. Each node corresponds to an item, and a directed edge $(u, v)$ indicates that item $u$ is a prerequisite for obtaining item $v$. Edge types specify the required action: \texttt{mine tool}, \texttt{craft}, or \texttt{place}. \texttt{craft station} edges indicate that crafting requires proximity to objects such as a table or furnace.

Given the target item and the agent's current inventory and observable state, the planner traverses the knowledge graph to identify unmet prerequisites and generates the corresponding subgoals. When a subgoal fails, the planner recomputes remaining steps from the updated state.

\subsection{Example Plan for Diamond Task from the LLM-Based Planner} \label{sec:high_level_plan_examples}

We present an example plan for the \texttt{Diamond} task generated by the LLM-based planner (\cref{lst:llm_full_context_init_plan}). The LLM-based planner tends to generate plans that resemble Minecraft conventions. For example, it treats the table and furnace as portable objects that can be placed and picked up, whereas in Crafter these are fixed structures. This results in suboptimal plans that require the low-level controller to track and revisit their locations. In contrast, the heuristic planner leverages the Crafter knowledge graph to generate plans consistent with the environment's mechanics.
\vspace{1em}

\begin{lstlisting}[caption=Initial Diamond Task Plan from LLM-Based Planner., label=lst:llm_full_context_init_plan]
mine('wood', None)
mine('wood', None)
mine('wood', None)
mine('wood', None)
mine('wood', None)
place('table')
craft('wood_pickaxe', {'wood': 1}, ['table'])
mine('stone', ['wood_pickaxe'])
mine('coal', ['wood_pickaxe'])
place('furnace')
craft('stone_pickaxe', {'wood': 1, 'stone': 1}, ['table'])
mine('iron', ['stone_pickaxe'])
craft('iron_pickaxe', {'wood': 1, 'coal': 1, 'iron': 1}, ['table', 'furnace'])
mine('diamond', ['iron_pickaxe'])
\end{lstlisting}

\subsection{Additional Planning Performance on Open-Weights LLMs} \label{sec:llm_planner_comparison}

\begin{table}[H]
\centering
\caption{Success rates of \dep~\citep{wang_describe_2023} with different open-weight LLMs as the high-level planner on three tasks that require multiple prerequisite steps.}
\begin{tabular}{cccc}
\toprule
\textbf{Model} & \textbf{Iron Pickaxe} & \textbf{Iron Sword} & \textbf{Diamond} \\ \midrule
gpt-oss-120b & 26\%                  & 32\%                & 5\%              \\
Qwen3.5-9B   & 37\%                  & 31\%                & 1\%              \\
Qwen3.5-27B  & 22\%                  & 41\%                & 7\%              \\
\bottomrule
\end{tabular}
\label{tbl:additional_llm_planner}
\end{table}

\clearpage
\section{Effect of Expert Training Seed Diversity} \label{sec:expert_seed_diversity}

\playdata is generated by one expert per reward, each from a single training run. Since one run converges to one particular policy, a natural concern is that the dataset covers only the behaviors of that policy. Because the Expert Behavior Generator trains experts from scratch, the number of training seeds is a controllable property of \playdata. We examine whether this axis adds useful variation beyond the two sources already built into \playdata: 20{,}000 procedurally generated worlds, one per episode, and reward-specific experts whose behaviors differ substantially (\cref{sec:cvpt_behavior_ablation}).

\textbf{Setup.} We train four independent experts for each of the Survival and Gathering rewards, which reach comparable performance across seeds (Expert Policy in \cref{tbl:crafter_scores}), and generate a second dataset with the same total number of episodes as \playdata. We then retrain \cvpt and \csteveone on each dataset and evaluate them as in \cref{sec:cvpt_eval,sec:csteveone_eval,sec:hierarchical_agent_eval}: Crafter Score and Return for \cvpt, and success rates on the six single-instruction tasks and the three long-horizon tasks for \csteveone.

\begin{table}[H]
\centering
\caption{\cvpt results with one or four expert-training seeds per reward.}
\small
\renewcommand{\arraystretch}{1.15}
\begin{tabular}{lccc}
\toprule
\textbf{Expert seeds} & \textbf{Model} & \textbf{Crafter Score (\%)} & \textbf{Return (\%)} \\
\midrule
\multirow{2}{*}{1 seed} & tiny & $\mathbf{95.6 \pm 1.6}$ & $\mathbf{93.2 \pm 1.4}$ \\
& base & $\mathbf{97.7 \pm 0.7}$ & $\mathbf{95.5 \pm 1.1}$ \\
\midrule
\multirow{2}{*}{4 seeds} & tiny & $93.4 \pm 3.3$ & $91.9 \pm 3.2$ \\
& base & $96.9 \pm 0.4$ & $95.1 \pm 0.7$ \\
\bottomrule
\end{tabular}
\label{tab:expert_seed_cvpt}
\end{table}

As shown in \cref{tab:expert_seed_cvpt}, using four expert-training seeds gives comparable \cvpt performance for both model sizes. For the current tasks, one seed per reward is sufficient to construct an effective behavioral-pretraining dataset, while additional seeds provide another way to vary the data.

\begin{table}[H]
\centering
\caption{\csteveone success rates with one or four expert seeds: short- and medium-horizon tasks.}
\small
\renewcommand{\arraystretch}{1.15}
\begin{tabular}{lrrrrrr}
\toprule
\textbf{Expert seeds} & \shortstack{\textbf{Place}\\\textbf{Table}} & \shortstack{\textbf{Place}\\\textbf{Furnace}} & \shortstack{\textbf{Make Wood}\\\textbf{Pickaxe}} & \shortstack{\textbf{Make Wood}\\\textbf{Sword}} & \shortstack{\textbf{Make Stone}\\\textbf{Pickaxe}} & \shortstack{\textbf{Make Stone}\\\textbf{Sword}} \\
\midrule
1 seed & 83\% & \textbf{100\%} & \textbf{98\%} & 82\% & 79\% & 85\% \\
4 seeds & \textbf{95\%} & 97\% & 84\% & \textbf{95\%} & \textbf{80\%} & \textbf{87\%} \\
\bottomrule
\end{tabular}
\label{tab:expert_seed_csteve_short}
\end{table}

\begin{table}[H]
\centering
\caption{\hdep + \csteveone success rates with one or four expert seeds: long-horizon tasks.}
\small
\renewcommand{\arraystretch}{1.15}
\begin{tabular}{lrrr}
\toprule
\textbf{Expert seeds} & \textbf{Iron Pickaxe} & \textbf{Iron Sword} & \textbf{Diamond} \\
\midrule
1 seed & 44\% & 46\% & \textbf{7\%} \\
4 seeds & \textbf{51\%} & \textbf{53\%} & 6\% \\
\bottomrule
\end{tabular}
\label{tab:expert_seed_csteve_long}
\end{table}

\Cref{tab:expert_seed_csteve_short,tab:expert_seed_csteve_long} show that four expert seeds change \csteveone success rates in both directions: gains on Place Table, Make Wood Sword, and the two iron-tool tasks are offset by a drop on Make Wood Pickaxe, and the remaining tasks change by at most three points. Together with \cref{tab:expert_seed_cvpt}, these results indicate that additional expert seeds do not substantially change downstream performance.

Overall, the world and objective diversity built into \playdata suffices for the current tasks, so a single expert seed per reward yields an effective dataset. The Expert Behavior Generator still exposes seed count, alongside reward design and dataset scale (\cref{sec:cvpt_scaling_ablation}), as a controlled factor.

\clearpage
\section{Asset Availability and Licenses} \label{sec:asset_availability}
\textbf{Released assets.} All assets introduced in this paper are publicly available.
\begin{itemize}
    \item \textbf{Code} (MIT License): \url{https://github.com/frechele/CrafterDojo}. The repository includes the training, evaluation, and data-generation scripts, dataset documentation, and Croissant metadata with Responsible AI fields for each dataset.
    \item \textbf{Model weights} (MIT License): \url{https://huggingface.co/frechele/CrafterDojo}, covering \cvpt (tiny, base, and large), \cclip, and \csteveone together with its prior.
    \item \textbf{Datasets} (CC0 1.0), hosted on Harvard Dataverse. Each dataset contains \playdata episodes together with their \clipcaption captions.
    \begin{itemize}
        \item CrafterPlay-Survival: \url{https://doi.org/10.7910/DVN/FHR2KS}
        \item CrafterPlay-Gatherer: \url{https://doi.org/10.7910/DVN/JXNHRB}
        \item CrafterDojo-sample: \url{https://doi.org/10.7910/DVN/KS1IAJ}
    \end{itemize}
\end{itemize}
\textbf{Existing assets.} We build on Craftax-Classic~\citep{matthews_craftax_2024} (MIT License). Our code adapts components of VPT~\citep{baker_video_2022}, MineCLIP~\citep{fan_minedojo_2022}, and CLIP~\citep{radford_learning_2021}, all released under the MIT License, and one utility file derived from JAX (Apache License 2.0). The \csteveone code is adapted from STEVE-1~\citep{lifshitz_steve-1_2023}, whose repository does not specify a license; rights to the original code remain with its authors. The LICENSE file of our repository reproduces all third-party license notices.

% \newpage
% \input{checklist.tex}

\end{document}